\documentclass{article}
\PassOptionsToPackage{numbers,square,comma,sort}{natbib}

\usepackage{amsmath}
\usepackage{bm}

 \usepackage[preprint]{neurips_2026}

\usepackage{caption}
\usepackage{graphicx}
\usepackage{booktabs}
\usepackage{neurips_2026}
\usepackage{colortbl}
\usepackage{multirow}
\usepackage{graphicx}
\usepackage{booktabs}
\usepackage[table]{xcolor}
\usepackage{graphicx}
\usepackage{adjustbox}
\usepackage{makecell}
\usepackage{array}
\usepackage[utf8]{inputenc} 
\usepackage[T1]{fontenc}    
\usepackage{xcolor}         
\definecolor{citeblue}{RGB}{0, 0, 255}
\usepackage{wrapfig}
\usepackage[
  colorlinks=true,
  linkcolor=black,
  citecolor=citeblue,
  urlcolor=citeblue
]{hyperref}       
\usepackage{url}            
\usepackage{booktabs}       
\usepackage{amsfonts}       
\usepackage{nicefrac}       
\usepackage{microtype}      

\let\cite\citep
\usepackage{url}            
\usepackage{booktabs}       
\usepackage{amsfonts}       
\usepackage{nicefrac}       
\usepackage{microtype}      

\title{\textbf{RoboStressBench: Benchmarking VLM Robustness to Physical Visual Stress in Embodied Scenes}
}

\author{
  \textmd{Leyi Wu\textsuperscript{1,3,$\ast$},~
  Yifan Zhao\textsuperscript{1,$\ast$},~
  Jinjie Zhang\textsuperscript{1,$\ast$},~
  Suzeyu Chen\textsuperscript{1,3,$\ast$},}\\
  \textmd{Wosong Chen\textsuperscript{1,3},~
  Zhifei Chen\textsuperscript{1},~
  Tianshuo Xu\textsuperscript{1},~
  Qingchun He\textsuperscript{1},}\\
  \textmd{Hongxin Hu\textsuperscript{1},~
  Haojian Huang\textsuperscript{1,3},~
  Yangkai Wei\textsuperscript{3},~
  Wenqian Li\textsuperscript{3},}\\
  \textmd{Yinchuan Li\textsuperscript{3},~
  Ying-Cong Chen\textsuperscript{1,2,$\dagger$}} \\
\textmd{\textsuperscript{1}HKUST(GZ) \hspace{2em} \textsuperscript{2}HKUST \hspace{2em} \textsuperscript{3}Knowin} \\
  \small\texttt{\ lwu398@connect.hkust-gz.edu.cn; yingcongchen@ust.hk}
}

\begin{document}

\maketitle
\maketitle
\makeatletter
\gdef\@fnsymbol#1{\ensuremath{\ifcase#1\or \ast \or \dagger \or \ddagger \or
  \mathsection \or \mathparagraph \or \| \or ** \or \dagger\dagger
  \or \ddagger\ddagger \else \fi}}
\makeatother
\let\thefootnote\relax\footnotetext{$\ast$ Equal contribution. Authors are listed in random order.}
\let\thefootnote\relax\footnotetext{$\dagger$ Corresponding author.}
\begin{center}
    \includegraphics[width=\textwidth]{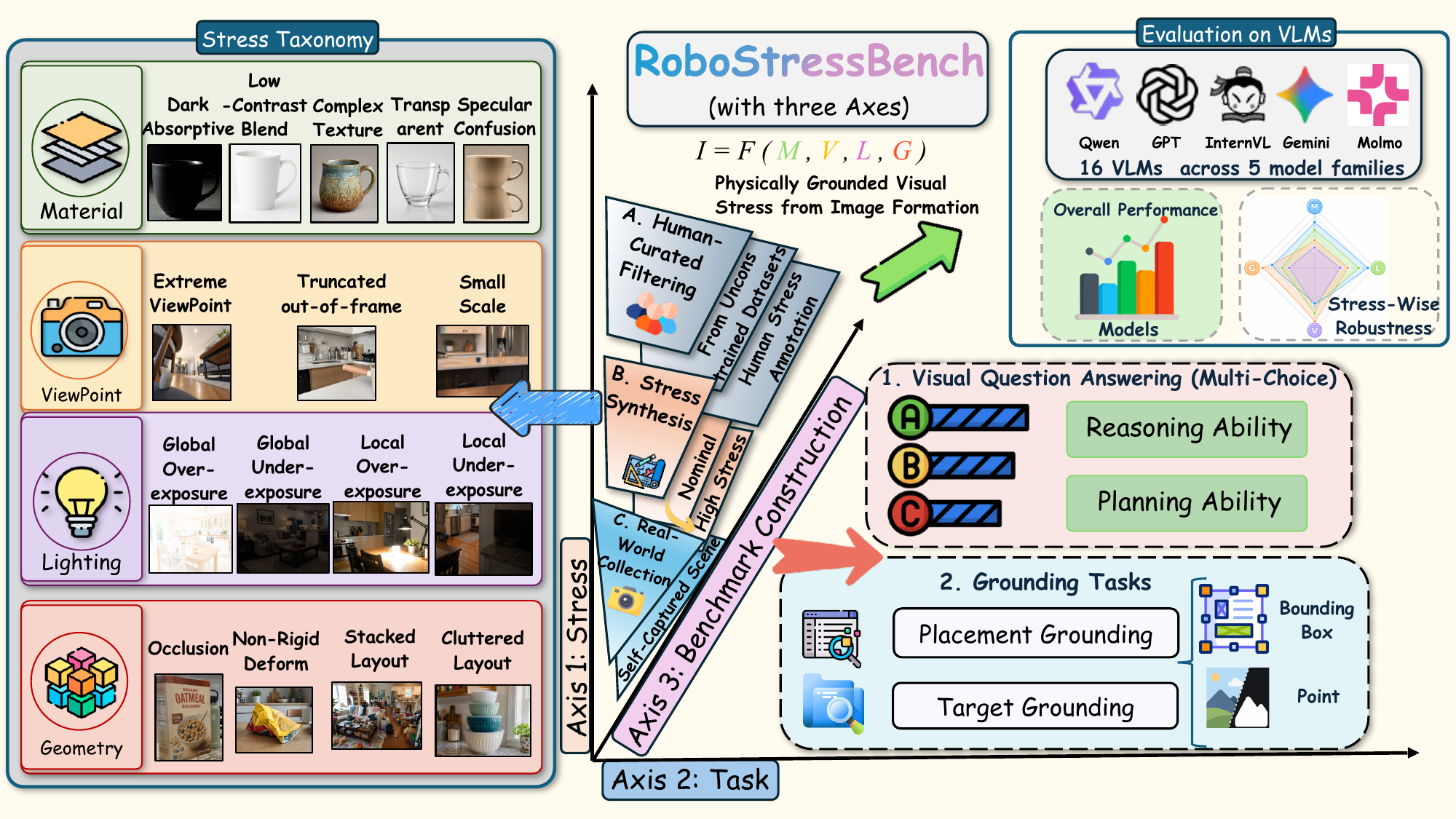}
    \captionof{figure}{
    \textbf{Overview of RoboStressBench.}
    RoboStressBench evaluates VLM robustness under physical visual stress in embodied scenes.
    We organize visual stress according to four image-formation factors: Material, Viewpoint, Lighting, and Geometry.
    The benchmark is constructed from human-curated filtering, stress synthesis, and real-world collection, and supports task-aligned evaluation through multiple-choice visual question answering and grounding tasks.
    We further evaluate diverse VLM families to analyze overall performance and stress-wise robustness.
    }
    \label{fig:overview}
\end{center}
\begin{abstract}
Vision-Language Models (VLMs) have shown strong visual understanding capabilities and are increasingly deployed in embodied AI systems, where reliable perception under real-world conditions is essential. 
However, existing benchmarks generally assess VLMs using clean images or isolated perturbations rather than stresses caused by physical scene formation. 
This design has two limitations: it covers only a narrow subset of everyday visual stresses, and some perturbations rarely appear in realistic embodied scenes.
This gap points to a more fundamental question: \textit{how can we define visual stress in a principled way that captures the diverse factors encountered in real physical environments?}
To address this question, we formulate visual perception from an \textit{inverse graphics} perspective and introduce \textbf{RoboStressBench}, a benchmark for systematically evaluating VLM robustness to physical visual stress in embodied scenes.
Inspired by the physical rendering equation, RoboStressBench decomposes visual stress into four physically grounded dimensions: Material ($M$), Viewpoint ($V$), Lighting ($L$), and Geometry ($G$). 
 This design enables RoboStressBench to cover a broad range of visual stresses that commonly arise in real-world environments, while allowing controlled analysis of their effects on VLM capabilities such as visual recognition, reasoning, and planning.
Through comprehensive evaluations of state-of-the-art VLMs, we identify stress-specific failure modes and reveal that different physical factors degrade different embodied capabilities, which are often obscured by aggregate accuracy.
We further introduce a stress-aware agentic solver that detects visual stressors and invokes visual-editing skills before reasoning, improving robustness in challenging high-stress scenarios. 
Overall, RoboStressBench provides a principled evaluation framework for diagnosing and improving VLM perception under real-world physical stress, supporting the development of more reliable embodied AI systems.
The project webpage is
~\textcolor{pink}{\href{https://yuevii.github.io/robostressbench-page/}{\texttt{RoboStressBench Page}}}.
\end{abstract}
\section{Introduction}
Recent Vision-Language Models (VLMs)~\cite{bai2025qwen3, qwen3.5, wang2025internvl3, molmo2openweightsdata, comanici2025gemini, achiam2023gpt} have achieved strong general visual understanding and zero-shot reasoning capabilities, making them increasingly attractive for embodied AI applications~\cite{liu2025hybridvla, kim2024openvla, driess2023palm, zitkovich2023rt}. 
However, for embodied agents to operate reliably in the real world, their visual perception must robustly handle a range of visual challenges. 
We refer to these challenges as \textit{physical visual stress}: visual degradation caused by physically plausible changes in scene appearance, where task-relevant evidence is weakened, distorted, or obscured. 
For example, a robot may need to recognize a transparent cup, localize a partially occluded tool, or make a decision under low illumination, specular reflection, or an unusual viewpoint. 
As shown in Table~\ref{tab:stress_aug_gap}, model accuracy drops on the same scene-question pairs after physically grounded stress editing, demonstrating the impact of physical visual stress on VLM reliability.

Existing benchmarks leave physical visual stress under-characterized in two ways
(see Fig.~\ref{fig:benchmark_comparison} for an overview).
General VLM benchmarks~\citep{yu2023mm, liu2024mmbench, li2023seed, fu2023mme} primarily evaluate broad abilities; visually challenging cases appear only incidentally and are rarely annotated with their underlying physical stress factors.
Robustness-oriented benchmarks~\cite{ishmam2025visual, li2026res, saxena2026vlm} explicitly evaluate degraded inputs, but often rely on ImageNet-C-style corruptions~\cite{hendrycks2019benchmarking}, such as noise, pixelation, or algorithmic blur. 
These digital perturbations are useful for robustness testing, but only partially reflect the physical visual stresses encountered in embodied scenes.
As a result, existing evaluations do not provide a principled way to diagnose how physical scene factors affect VLM reliability.

To address this gap, we introduce \textbf{RoboStressBench}, a benchmark for evaluating VLM robustness under physically grounded visual stress in embodied scenes. 
Inspired by inverse graphics, we abstract image formation as $I=\mathcal{F}(M,V,L,G)$ and organize stress into four dimensions: \textbf{Material} ($M$), \textbf{Viewpoint} ($V$), \textbf{Lighting} ($L$), and \textbf{Geometry} ($G$). 
These dimensions provide an interpretable framework for diagnosing whether failures arise from surface appearance, camera pose, illumination, or spatial structure.
We construct RoboStressBench through three complementary sources: \textbf{filtering}, \textbf{synthesis}, and \textbf{collection}. 
We filter naturally occurring stress cases from existing datasets, synthesize targeted stress variants from nominal images for rare or hard-to-isolate categories, and collect additional real-world examples from Internet-sourced and self-captured images. 
This pipeline balances natural realism, stress diversity, and factor-level controllability.

Using RoboStressBench, we evaluate 16 state-of-the-art VLMs across five model families, including Qwen~\cite{qwen3.5}, InternVL~\cite{wang2025internvl3}, Molmo~\cite{molmo2openweightsdata}, GPT~\cite{achiam2023gpt}, and Gemini~\cite{comanici2025gemini}. 
Our results show that physical visual stress affects models unevenly: geometry stress strongly degrades localization and spatial reasoning, while material and lighting stress more often affect recognition and state understanding. 
These task-stress interactions reveal failure modes that are hidden by aggregate accuracy, motivating stress-aware evaluation beyond a single overall score.

As a proof-of-concept intervention enabled by this diagnosis, we further introduce \textbf{StressDART}, a stress-aware test-time solver that detects the dominant stress factor, applies targeted visual rectification, and reasons over the original and rectified images. 
StressDART yields modest robustness gains without model fine-tuning, suggesting that explicit stress diagnosis can guide test-time interventions while also highlighting the need for content-preserving rectification.

In summary, our contributions are as follows:

$\bullet$ We introduce \textbf{RoboStressBench}, a benchmark and evaluation protocol for diagnosing VLM robustness in embodied scenes. 
RoboStressBench provides a physically grounded way to characterize visual difficulty, covering common real-world stressors caused by material appearance, camera viewpoint, illumination, and scene geometry.

$\bullet$ We construct an approximately \textbf{7.2K} visual stress dataset through human-annotated filtering, controlled synthesis, and real-world data collection, balancing realism, diversity, and controllability.

$\bullet$ We provide a systematic diagnostic analysis of VLM robustness under physical visual stress, revealing task-stress interactions and stress-specific failure modes that are obscured by aggregate accuracy.

$\bullet$ We propose \textbf{StressDART}, a modular stress-aware agentic solver that detects visual stress, applies targeted visual rectification, and performs reasoning on the processed input. 
Experiments show that explicit stress diagnosis improves robustness under challenging physical conditions.

\newcommand{\figheight}{0.28\textheight}
\begin{figure}[t]
\centering

\begin{minipage}[t]{0.48\linewidth}
\centering
\vspace{0pt}
\includegraphics[height=\figheight,width=\linewidth,keepaspectratio]{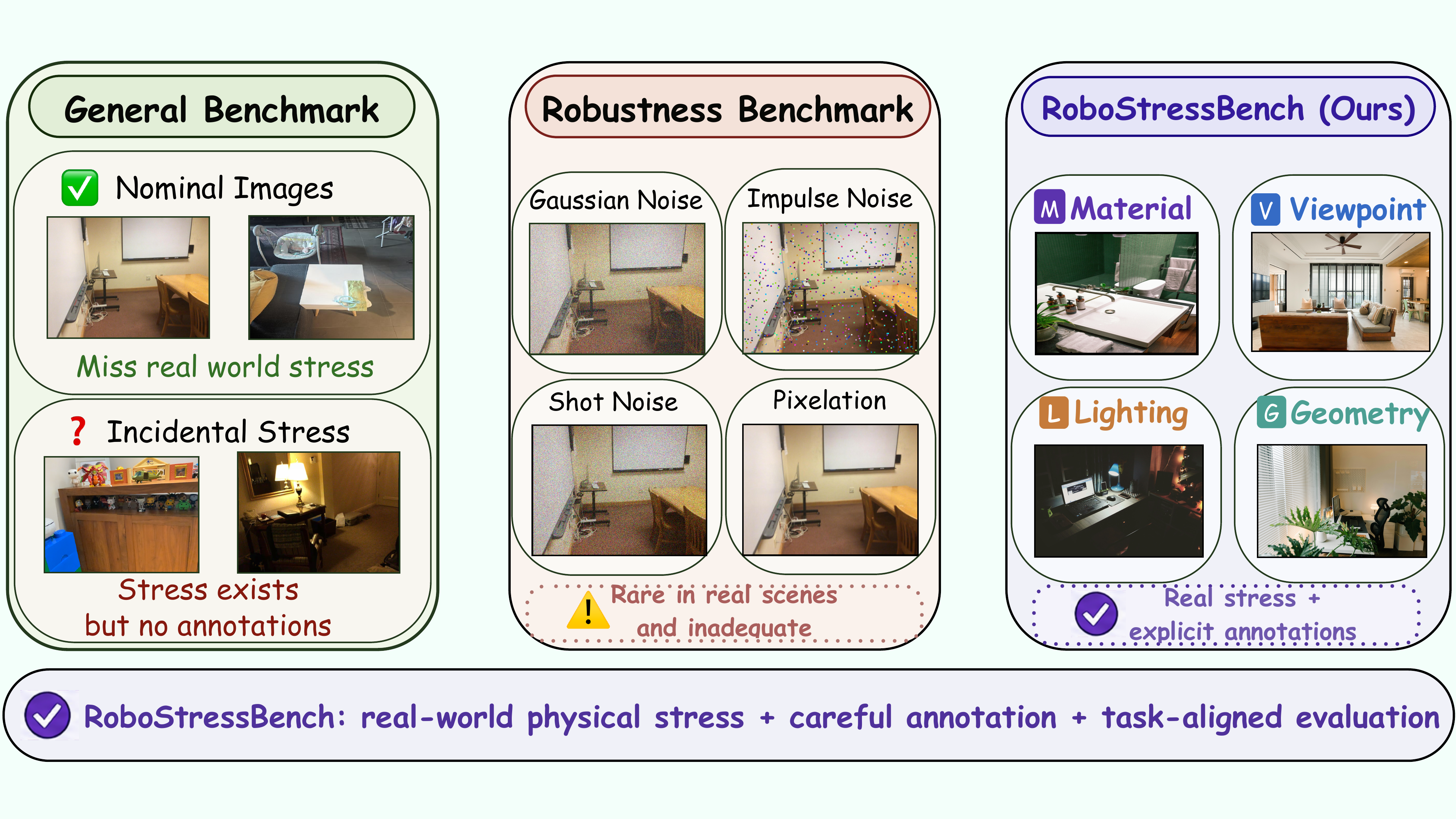}
\captionof{figure}{
\textbf{Motivation for RoboStressBench.}
Existing benchmarks either lack explicit stress annotation or rely on artificial perturbations, whereas RoboStressBench provides realistic physical stress with careful annotations.
}
\label{fig:benchmark_comparison}
\end{minipage}
\hfill
\begin{minipage}[t]{0.48\linewidth}
\centering
\vspace{0pt}
\includegraphics[height=\figheight,width=\linewidth,keepaspectratio]{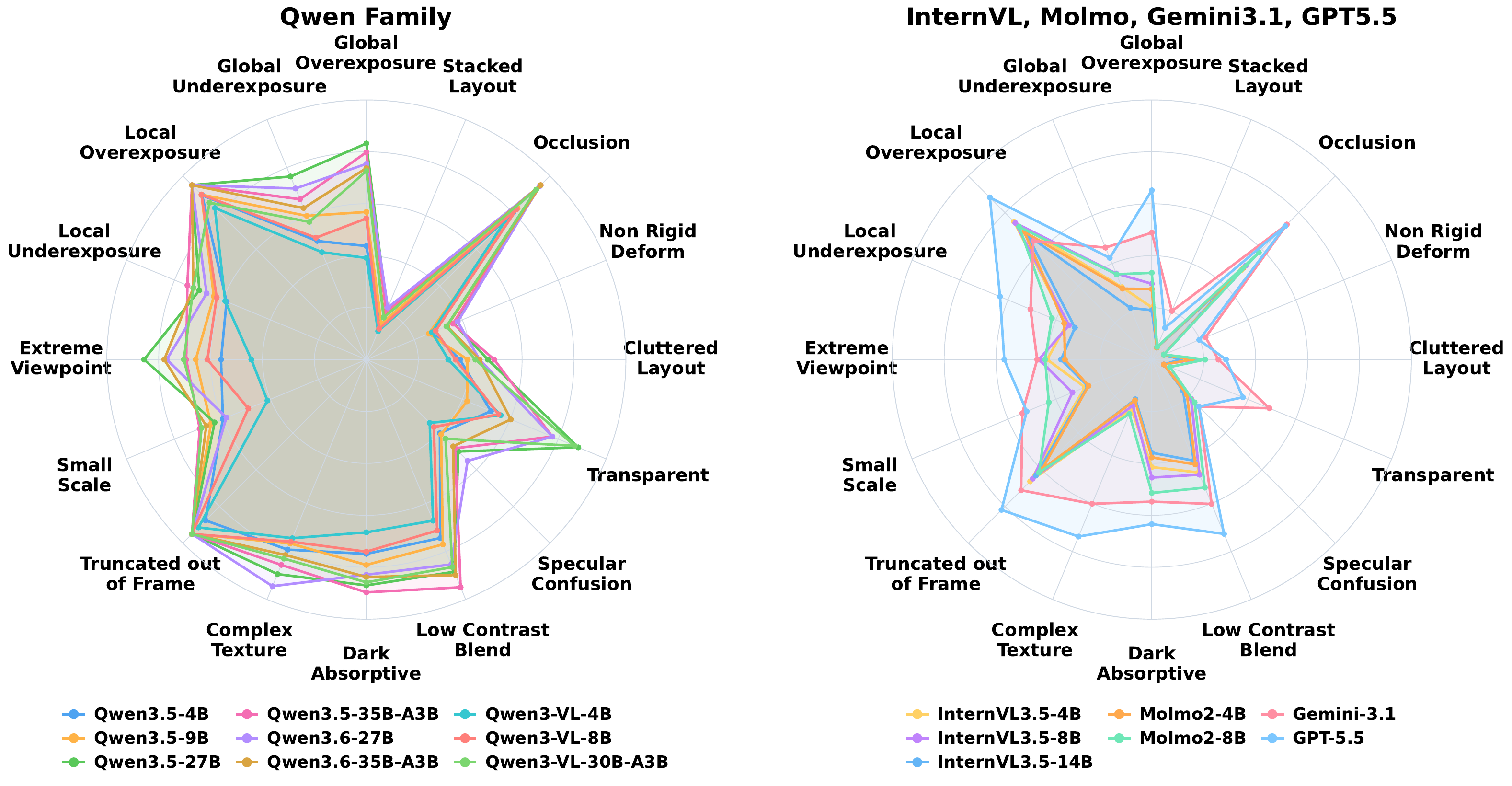}
\captionof{figure}{
\textbf{RoboStressBench evaluation results.}
We visualize the performance of all evaluated VLMs across RoboStressBench stress dimensions. 
Comprehensive numerical results are reported in Table~\ref{tab:model_comparison}.
}
\label{fig:radar_large}
\end{minipage}

\vspace{-5mm}
\end{figure}

\section{Related Work}

\paragraph{From Visual Corruption to Physical Stress.}
Investigating how visual inputs challenge model robustness has motivated extensive research into stress and perturbations. 
Early work characterized visual vulnerability through worst-case perturbations
\begin{wraptable}{r}{0.48\textwidth}
    \centering
    \caption{
    \textbf{Effect of physical visual stress on the paired editing subset.}
    We compare VLM accuracy on the same scene-question pairs before and after stress editing, showing the impact of physically grounded stress on model performance.
    }

    \label{tab:stress_aug_gap}
    {\setlength{\tabcolsep}{3pt}
    \renewcommand{\arraystretch}{1.05}
    \resizebox{0.36\textwidth}{!}{
    \tiny
    \begin{tabular}{lccc}
    \toprule
    \multirow{2}{*}{\textbf{Family}} & \multicolumn{3}{c}{Accuracy (\%)} \\
    \cmidrule(lr){2-4}
           & \textbf{Nom.} & \textbf{Stress} & \textbf{Drop} \\
    \midrule
    \textbf{Qwen3VL} & 51.0 & 35.5 & -15.5 \\
    \textbf{Qwen3.5} & 53.5 & 36.8 & -16.8 \\
    \textbf{Qwen3.6} & 64.3 & 40.1 & -24.1 \\
    \textbf{InternVL3.5} & 10.0 & 9.9 & -0.1 \\
    \textbf{Molmo2} & 12.2 & 11.5 & -0.7 \\
    \bottomrule
    \end{tabular}
    }}
\end{wraptable}
~\citep{szegedy2013intriguing,goodfellow2014explaining,madry2017towards}. ImageNet-C/P extended this view to non-adversarial corruptions, organizing stress into controllable families
~\citep{hendrycks2019benchmarking}.
Another line of work studies natural distribution shifts that better reflect deployment conditions. These benchmarks cover real-world shifts such as background, and rotation, as well as hard natural images, rendition/sketch shifts~\citep{barbu2019objectnet,koh2021wilds}. More recent efforts, such as ImageNet-3DCC~\citep{kar20223d}
and ImageNet-D~\citep{zhang2024imagenet}, move toward physically plausible or high-level controllable stress.
However, existing taxonomies often focus on isolated stress families and lack a unified physical account. RoboStressBench addresses this by grounding visual stress in image formation and provides interpretable dimensions for benchmarking, failure attribution, and stress-aware VLM reasoning.

\paragraph{Robustness Evaluation for Multimodal Understanding.}

Robustness benchmarking has evolved from image classification to increasingly complex perception tasks. In classification, ImageNet-C/P~\citep{hendrycks2019benchmarking} established a standard protocol to measure the models' robustness. This protocol was later extended to object detection
~\citep{michaelis2019benchmarking} and semantic segmentation robustness~\citep{kamann2020benchmarking}. 
Recent multimodal benchmarks have made robustness evaluation more relevant to VLMs. The Visual Robustness Benchmark for VQA~\citep{ishmam2025visual} evaluates VQA models and MLLMs under realistic visual corruptions with robustness-oriented metrics.
Res-Bench~\citep{li2026res} focuses on MLLM resolution robustness, measuring performance stability and volatility across dynamic input resolutions.
VLM-RobustBench directly evaluates VLMs under a wide range of augmentations across visually grounded and reasoning-oriented datasets~\citep{saxena2026vlm}. Some works such as R-Bench~\citep{li2025r}, Eva-VLA~\citep{liu2025eva} and DarkEQA~\citep{park2025darkeqa}, study multimodal robustness under real-world corruptions and physical variations.
However, existing benchmarks rarely diagnose VLM failures through the physical image-formation factors. RoboStressBench fills this gap by evaluating VLMs along four interpretable stress dimensions.


\paragraph{Embodied Benchmarks for Vision-Language Models.}
Embodied VLM evaluation has gone beyond image QA, 
evolving from testing \textit{whether embodied agents can answer questions} to evaluating \textit{whether visual evidence can guide what to localize, how to reason spatially, and where to act}.
OpenEQA~\cite{majumdar2024openeqa} and RoboVQA~\cite{sermanet2024robovqa} exemplify this question-answering paradigm, testing situated understanding over scene observations, visual memory, task progress, and robot experience.

Closer to action, the RoboRefIt~\cite{lu2023vlgrasp} dataset supports grounding language to manipulable objects and grasp targets, while RoboSpatial~\cite{song2025robospatial} and RefSpatial-Bench~\cite{zhou2025roborefer} extend evaluation to robotics-oriented 2D/3D spatial reasoning and multi-step referring in robot-centered scenes.
More action-centric benchmarks~\citep{yuan2025robopoint,hao2025roboaffordpp,yuan2026seeing} evaluate decision-relevant visual outputs
In parallel, broader evaluations~\citep{chen2025robo2vlm,qiu2024egoplanbench2,yang2025thinking,yang2025embodiedbench} extend this trajectory along temporal, memory, planning, and agent-level dimensions.

However, task-level scores often conflate perception, reasoning, and planning errors. RoboStressBench complements them by diagnosing failures along physical image-formation axes.


\section{Preliminaries}
\label{sec:preliminaries}

\paragraph{Image Formation and Visual Stress.}
We use physically based rendering as a conceptual basis for defining physical visual stress. 
The rendering equation~\cite{kajiya1986rendering} models the outgoing radiance at a surface point $\mathbf{x}$ along direction $\boldsymbol{\omega}_o$ as
\begin{equation}\label{eq:rendering}
L_o(\mathbf{x}, \boldsymbol{\omega}_o)
=
\int_{\Omega}
f_r(\mathbf{x}, \boldsymbol{\omega}_i, \boldsymbol{\omega}_o)
L_i(\mathbf{x}, \boldsymbol{\omega}_i)
\max(0, \boldsymbol{\omega}_i \cdot \mathbf{n})
d\boldsymbol{\omega}_i ,
\end{equation}
where $f_r$ is the Bidirectional Reflectance Distribution Function (BRDF), $L_i$ denotes incident radiance from direction $\boldsymbol{\omega}_i$, and $\mathbf{n}$ is the surface normal at $\mathbf{x}$. 
Although this equation is not a complete camera model, it highlights several physical factors that shape image appearance, including material reflectance, illumination, viewing direction, and surface geometry. 
Following the inverse graphics perspective, we abstract image formation as
\begin{equation}
I = \mathcal{F}(M, V, L, G),
\end{equation}
where $M$, $V$, $L$, and $G$ denote \textbf{Material}, \textbf{Viewpoint}, \textbf{Lighting}, and \textbf{Geometry}, respectively. 
These factors correspond to interpretable components of image formation: $M$ is associated with reflectance properties such as $f_r$, $L$ with incident illumination such as $L_i$, $V$ with viewing direction $\boldsymbol{\omega}_o$, and $G$ with spatial structure such as surface position and normal $(\mathbf{x}, \mathbf{n})$. 
We define \textit{physical visual stress} as physically plausible states of these factors that make task-relevant visual evidence less accessible to VLMs while leaving the underlying scene semantics unchanged. 
RoboStressBench instantiates this abstraction as material, viewpoint, lighting, and geometry stress, covering phenomena such as transparency, low illumination, unusual camera poses, occlusion, and clutter.

\begin{figure}[t]
    \centering
    \includegraphics[width=\linewidth]{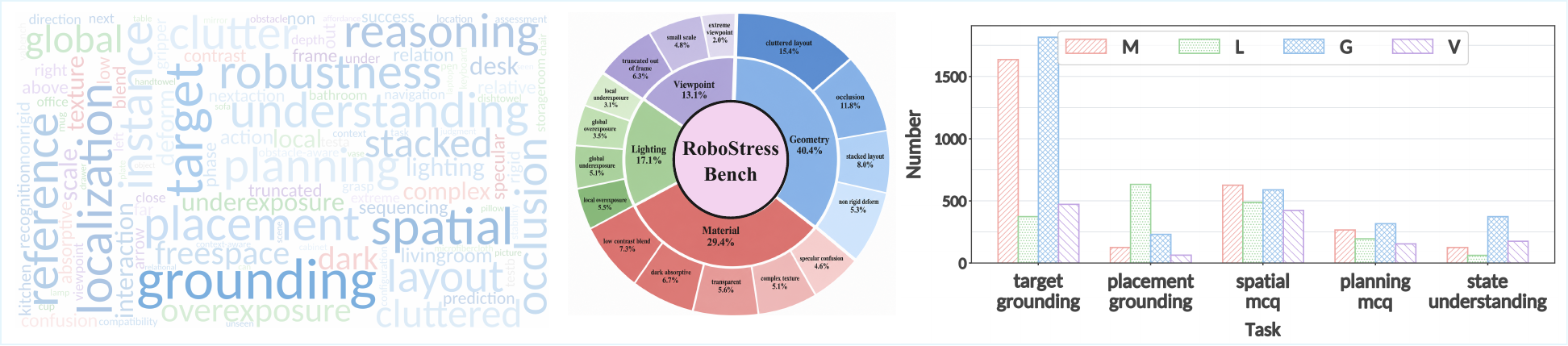}
    \caption{Overview of \textbf{RoboStressBench}’s statistical distributions. (\textbf{\textit{Left}}) Word distribution of prompt suites; (\textbf{\textit{Middle}}) Data distribution across 16 sub-stress types; and (\textbf{\textit{Right}}) Data distribution across different tasks.
    }
    \label{fig:statistics}

\end{figure}
\section{RoboStressBench: Benchmarking Physical Visual Stress in Embodied Scenes}
\label{sec:benchmark}
\begin{figure}[t]
    \centering
    \includegraphics[width=0.9\linewidth]{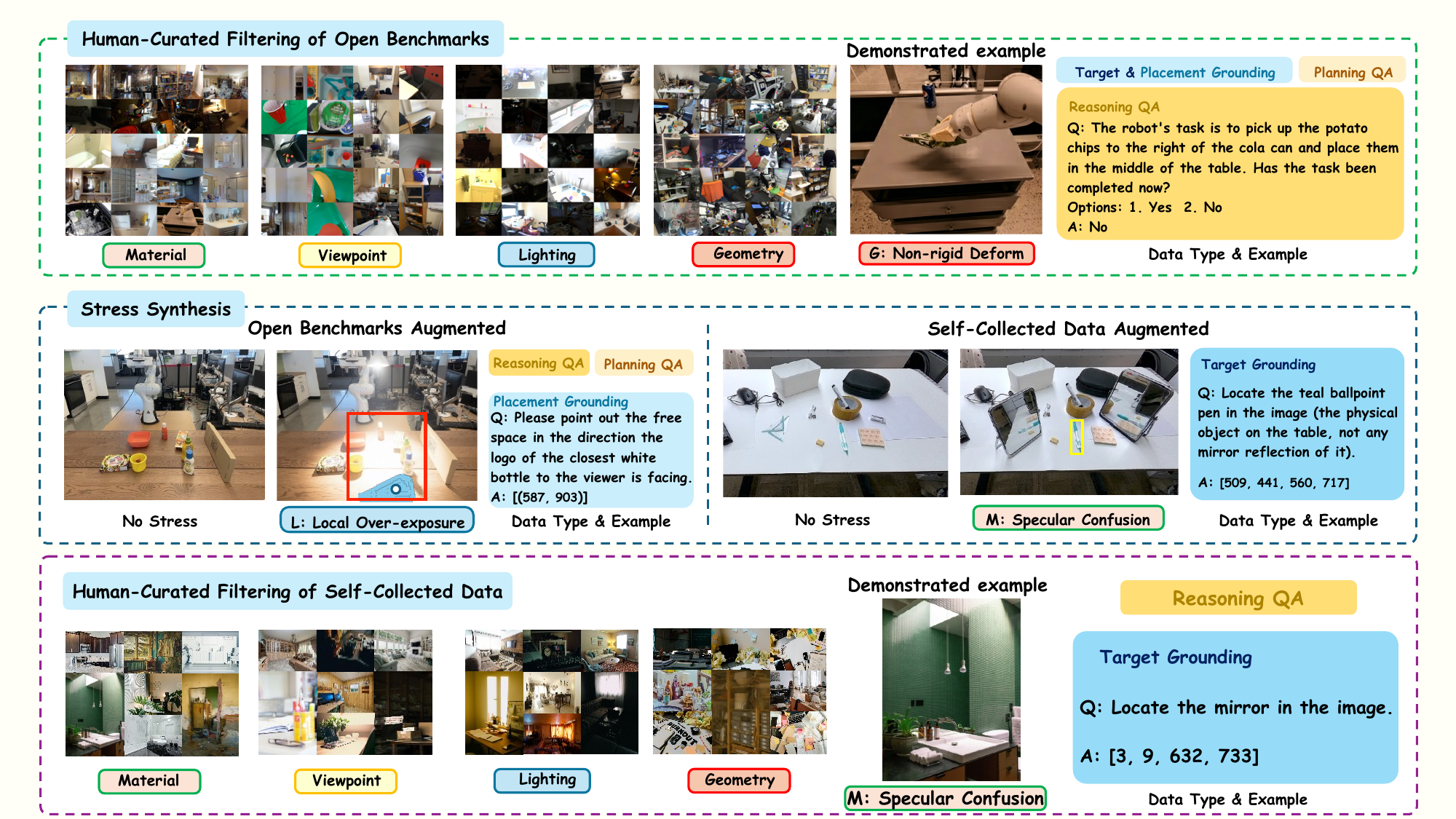}
    \caption{
    \textbf{Stress categories and curation pipeline.}
    Overview of the four stress dimensions and three data sources in RoboStressBench.
    }
    \label{fig:stress}

\end{figure}

We first introduce the stress taxonomy (Sec.~\ref{sec:taxonomy}) and then describe the dataset curation pipeline (Sec.~\ref{sec:curation}). Fig.~\ref{fig:overview} provides an overview of RoboStressBench, Fig.~\ref{fig:statistics} summarizes the dataset statistics, and Fig.~\ref{fig:stress} illustrates the stress categories and the overall curation pipeline.

\subsection{Stress Taxonomy}
\label{sec:taxonomy}
RoboStressBench organizes visual stress using a physically grounded taxonomy based on the image formation abstraction $I=\mathcal{F}(M,V,L,G)$. 
We define four primary stress dimensions: \textbf{Material ($M$)}, \textbf{Viewpoint ($V$)}, \textbf{Lighting ($L$)}, and \textbf{Geometry ($G$)}. 
Each dimension is further divided into fine-grained stress categories for controlled diagnosis of VLM perception and reasoning.

\paragraph{Material Stress.}
Material stress arises from surface appearance properties that obscure object identity, boundaries, or semantic cues. 
We consider five material-related stress types: \textit{dark absorptive}, where objects or surfaces absorb most incident light and lose visible detail; \textit{low-contrast blend}, where the target visually blends into the background due to similar color, texture, or brightness; \textit{complex texture}, where highly patterned surfaces interfere with recognition; \textit{transparent}, where refraction or background visibility changes object appearance; and \textit{specular confusion}, where mirror-like or glossy reflections introduce misleading visual evidence.

\paragraph{Viewpoint Stress.}
Viewpoint stress is caused by camera pose, object scale, or framing conditions that make an object depart from its canonical appearance. 
We define three viewpoint-related stress types: \textit{extreme viewpoint}, covering unusual observation angles such as top-down, low-angle, or side views; \textit{truncated out-of-frame}, where the target is partially outside the image boundary; and \textit{small scale}, where the target occupies only a small image region and becomes difficult to recognize or localize.

\paragraph{Lighting Stress.}
Lighting stress is caused by illumination conditions that suppress, saturate, or unevenly distort visual evidence. 
We define four lighting-related stress types: \textit{global overexposure}, where excessive illumination washes out most of the scene; \textit{local overexposure}, where strong light, glare, or highlights saturate specific regions; \textit{global underexposure}, where the entire scene is too dark to reveal sufficient detail; and \textit{local underexposure}, where shadows or uneven lighting obscure only part of the image.
\paragraph{Geometry Stress.}
Geometry stress arises from spatial structure, deformation, occlusion, and object arrangement. 
We consider four geometry-related stress types: \textit{occlusion}, where the target is partially blocked by another object or scene element; \textit{non-rigid deform}, where object shape changes due to bending, folding, compression, or related transformations; \textit{stacked layout}, where objects are piled or layered vertically and support relations become ambiguous; and \textit{cluttered layout}, where dense object arrangements make segmentation and spatial reasoning difficult.

This taxonomy enables two-level diagnosis: dimension-level analysis across Material, Viewpoint, Lighting, and Geometry, and category-level analysis within each dimension. 
As a result, RoboStressBench can identify not only whether a model fails under stress, but also which physical factor and fine-grained stress pattern are associated with the failure.

\subsection{Dataset Curation}
\label{sec:curation}
RoboStressBench is curated from three complementary sources to balance realism, diversity, and controllability. 
First, we select naturally occurring stress cases from existing unconstrained datasets~\cite{du2024embspatial, zhou2025roborefer, 10.1145/3746027.3758209, song2025robospatial, chen2025robo2vlm, yuan2026seeing, yuan2025robopoint, lu2023vlgrasp}. 
Second, we synthesize targeted stress variants from nominal images for categories that are rare or difficult to isolate in real data. 
Third, we collect additional real-world examples from Internet-sourced and self-captured images.
Since physical stress factors often co-occur in real scenes, RoboStressBench supports multi-label stress annotation and records the dominant stress dimension for factor-level analysis.

RoboStressBench supports both visual question answering (VQA) and grounding tasks. 
We retain original questions or grounding annotations when available and manually verified; otherwise, annotators create task-specific questions, answers, or grounding labels. 
For synthesized grounding examples, we transfer annotations when the nominal and stressed images remain pixel-aligned after resizing, and re-label the target region otherwise. 
Detailed dataset statistics and annotation protocols are provided in Appendix~\ref{supp:details}.
All examples and annotations are provided in the supplementary material. 

\section{StressDART: Test-Time Stress Detection and Rectification for Robust Visual Reasoning}
\begin{figure}[t]
    \centering
    \includegraphics[width=0.5\linewidth]{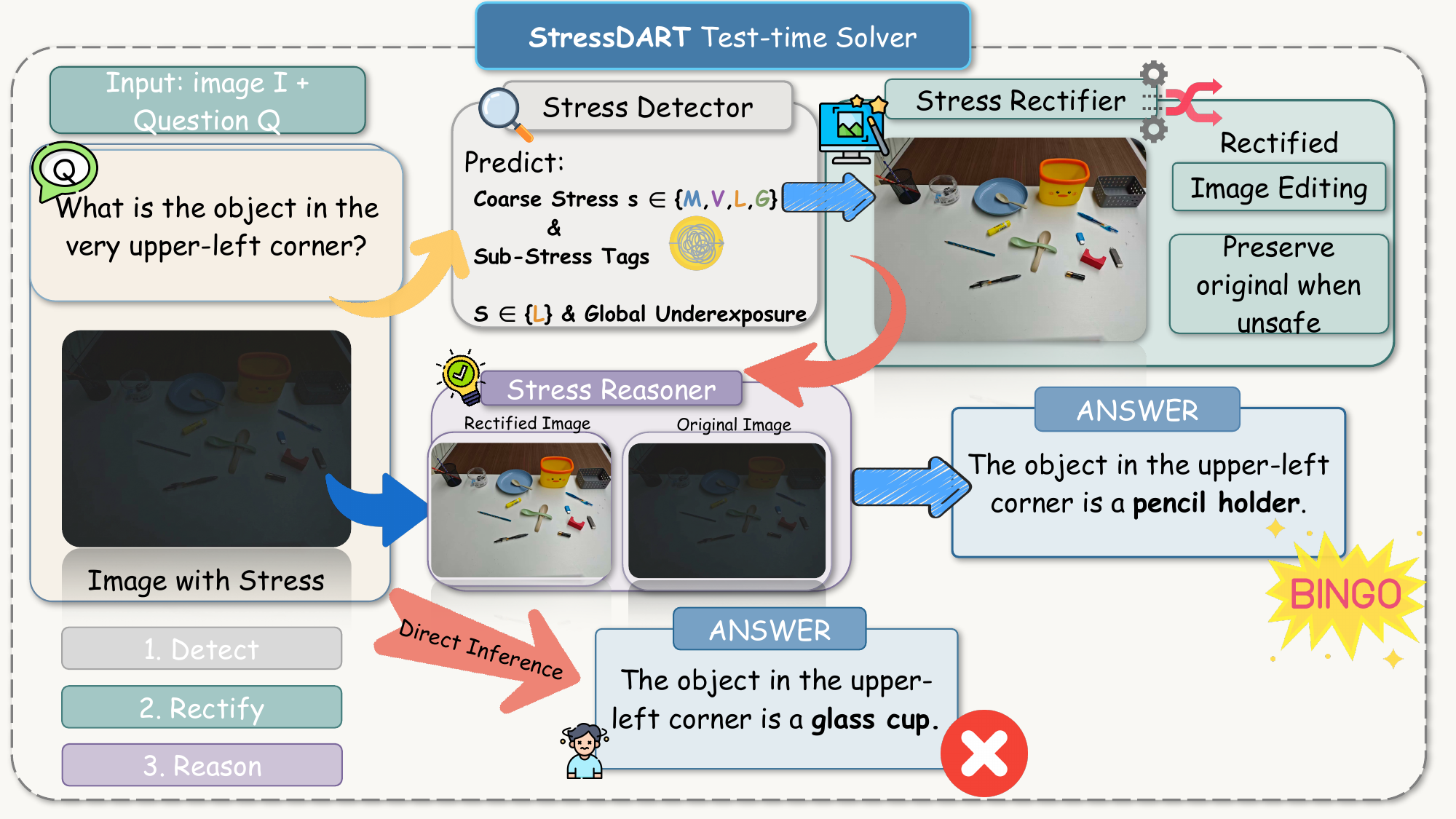}
    \caption{
    \textbf{Overview of StressDART.}
    Given a stressed image and a question, StressDART first detects the dominant visual stress, then applies targeted rectification to recover task-relevant evidence, and finally reasons over both the original and rectified images to produce the answer.
    }
    \label{fig:stressdart}

\end{figure}

RoboStressBench reveals that many VLM failures under physical visual stress are tied to identifiable scene factors, such as poor illumination, specular surfaces, occlusion, or unusual viewpoints. 
This motivates a test-time strategy that first diagnoses the dominant stressor and then applies a targeted operation to recover task-relevant visual evidence. 
We therefore propose \textbf{StressDART}, a stress-aware solver for \textbf{D}etection \textbf{A}nd \textbf{R}ectification at \textbf{T}est time. 
As shown in Fig.~\ref{fig:stressdart}, StressDART requires no model fine-tuning and consists of three stages: stress detection, stress rectification, and final reasoning.

Given an image $I$ and a question $Q$, StressDART first uses a \textbf{Stress Detector} to predict the stress condition relevant to the task:
\begin{equation}
s, c = \mathcal{D}(I, Q),
\end{equation}
where $s \in \{M,V,L,G\}$ denotes the coarse stress dimension and $c$ denotes a fine-grained stress category, such as \textit{transparent}, \textit{global underexposure}, \textit{occlusion}, or \textit{small scale}. 
This explicit diagnosis allows subsequent processing to be conditioned on why the image is difficult.

Next, a \textbf{Stress Rectifier} selects a category-specific visual operation $\phi_c$ and applies it to the input image:
\begin{equation}
\tilde{I} = \phi_c(I),
\end{equation}
where $\tilde{I}$ is the rectified image. 
For example, underexposure may trigger illumination enhancement, overexposure may trigger highlight recovery, and small-scale targets may trigger cropping or zooming. 
For stressors that cannot be safely corrected, the rectifier preserves the original image or applies only conservative transformations.

Finally, the \textbf{Reasoner} answers the original question using both the original and rectified visual evidence:
\begin{equation}
A = \mathcal{R}(I, \tilde{I}, Q, s, c),
\end{equation}
where $\mathcal{R}$ is the VLM reasoner and $A$ is the predicted answer. 
Providing both $I$ and $\tilde{I}$ preserves the original task context while allowing the model to exploit recovered visual cues. 
By separating diagnosis, rectification, and reasoning, StressDART provides an interpretable test-time framework for improving VLM robustness under physical visual stress.

\section{Experiments}
\label{sec:exp}
We evaluate RoboStressBench from three complementary perspectives. 
First, we benchmark a broad set of state-of-the-art VLMs to characterize their robustness under physical visual stress. 
Second, we analyze performance across Material, Viewpoint, Lighting, and Geometry to identify how different image-formation factors affect performance. 
Third, we evaluate StressDART to test whether explicit stress diagnosis and targeted visual rectification can improve robustness at test time.

\subsection{Experimental Settings}

\paragraph{Evaluation Protocol.}
We evaluate models on multiple-choice and grounding tasks. 
For multiple-choice questions, we report exact-match accuracy over the predicted option. 
For grounding tasks, we evaluate point predictions by checking whether the point falls inside the ground-truth mask, and evaluate box predictions using IoU-based metrics. 
The grounding scores in Table~\ref{tab:model_comparison} average point-based grounding accuracy and box-based IoU@0.95. 
Additional grounding metrics and evaluation details are provided in Appendix~\ref{supp:grounding_metrics}.

\paragraph{Models.}
We evaluate a broad collection of open-source and closed-source VLMs. 
For open-source models, we include representative families: \textbf{Qwen3-VL}~\cite{bai2025qwen3} with 4B, 8B, and 30B-A3B variants;
\textbf{Qwen3.5}~\cite{qwen3.5} with 4B, 9B, 27B, and 35B-A3B variants; \textbf{Qwen3.6}~\cite{qwen3.6-35b-a3b} with 27B and 35B-A3B variants; \textbf{InternVL3.5}~\cite{wang2025internvl3} with 4B, 8B, and 14B variants; and \textbf{Molmo2}~\cite{molmo2openweightsdata} with 4B and 8B variants. 
For commercial models, we evaluate \textbf{Gemini-3.1}~\cite{comanici2025gemini} and \textbf{GPT-5.5}~\cite{achiam2023gpt}. 
In total, our evaluation covers 16 VLMs across 5 model families. 
In StressDART, we use Qwen3-VL-4B~\cite{bai2025qwen3} as both the Stress Detector and the final Reasoner, and instantiate the Stress Rectifier with Qwen-Image-Edit~\cite{wu2025qwenimagetechnicalreport} to produce rectified visual inputs at test time.

\paragraph{Implementation Details.}
For open-source models, we use their official inference pipelines with deterministic greedy decoding, setting the maximum generation length to 64 new tokens and disabling sampling (temperature $=0.0$, top-$p=1.0$).
For commercial models, we query the official APIs using the same image-question format and the same generation budget (maximum output tokens $=64$, temperature $=0.0$, top-$p=1.0$).
All models are evaluated with a unified instruction template and are constrained to produce answers in the required format. 

\definecolor{first}{RGB}{247,176,103}
\definecolor{second}{RGB}{250,210,122}
\definecolor{third}{RGB}{252,237,170}

\newcommand{\best}[1]{\cellcolor{first}\textbf{#1}}
\newcommand{\secondbest}[1]{\cellcolor{second}\textbf{#1}}
\newcommand{\thirdbest}[1]{\cellcolor{third}\textbf{#1}}

\begin{table*}[t]
\centering

{\setlength{\tabcolsep}{3pt}
\renewcommand{\arraystretch}{1.2}
\scriptsize
\resizebox{\textwidth}{!}{
\begin{tabular}{ll|c|cccccccccccccccc|ccccc}
\toprule
\multirow{3}{*}{\textbf{Model}} & \multirow{3}{*}{\textbf{Size}} & \multirow{3}{*}{\textbf{Overall}} & \multicolumn{16}{c|}{\textbf{Stress Dimensions}} & \multicolumn{5}{c}{\textbf{Task Dimensions}} \\
\cmidrule(lr){4-19} \cmidrule(lr){20-24}
 &  &  & \multicolumn{5}{c}{\textbf{Material}} & \multicolumn{3}{c}{\textbf{Viewpoint}} & \multicolumn{4}{c}{\textbf{Geometry}} & \multicolumn{4}{c|}{\textbf{Lighting}} & \multicolumn{2}{c}{\textbf{Grounding}} & \multicolumn{2}{c}{\textbf{Reasoning}} & \multicolumn{1}{c}{\textbf{Planning}} \\
\cmidrule(lr){4-8} \cmidrule(lr){9-11} \cmidrule(lr){12-15} \cmidrule(lr){16-19} \cmidrule(lr){20-21} \cmidrule(lr){22-23} \cmidrule(lr){24-24}
 &  &  & \textbf{Dark} & \textbf{L-Con} & \textbf{C-Tex} & \textbf{Tran.} & \textbf{Spec.} & \textbf{Extr.} & \textbf{Trun.} & \textbf{Small} & \textbf{Occl.} & \textbf{Non-R} & \textbf{Stack} & \textbf{Clust} & \textbf{G-Ovr} & \textbf{L-Ovr} & \textbf{G-Und} & \textbf{L-Und} & \textbf{Plc.} & \textbf{Tgt.} & \textbf{Spa.} & \textbf{Sta.} & \textbf{Plan} \\
\midrule

\textbf{Qwen3VL} & 4B & 43.2 & 50.6 & 49.8 & 53.7 & 44.0 & 32.4 & 38.1 & 62.0 & 36.6 & 53.4 & 27.9 & 16.8 & 30.6 & 34.5 & 57.7 & 38.4 & 45.6 & 34.1 & 31.9 & 65.2 & 49.4 & 53.8 \\
\textbf{Qwen3VL} & 8B & 49.7 & 58.9 & 57.6 & 59.9 & 49.0 & 38.4 & 52.4 & 69.8 & 45.8 & 62.8 & 33.3 & 21.5 & 36.9 & 48.6 & 66.6 & 46.6 & 53.0 & 45.3 & 36.2 & 73.4 & 58.8 & 64.2 \\
\textbf{Qwen3VL} & 30B-A3B & 55.9 & \thirdbest{64.7} & 65.0 & 63.3 & \secondbest{65.2} & 42.2 & 57.1 & \thirdbest{70.8} & \secondbest{56.2} & 67.9 & 36.3 & 25.2 & 41.6 & 58.2 & 64.5 & 50.3 & \secondbest{58.0} & 39.4 & 41.1 & 71.9 & \best{68.6} & \best{99.8} \\
\midrule
\textbf{Qwen3.5} & 4B & 49.8 & 59.4 & 59.1 & 61.5 & 47.3 & 40.4 & 49.5 & 65.6 & 51.6 & 61.9 & 32.3 & 22.4 & 37.9 & 42.6 & 66.4 & 45.9 & 50.8 & 39.4 & 37.1 & 72.6 & 59.6 & 68.8 \\
\textbf{Qwen3.5} & 9B & 50.7 & 61.5 & 60.4 & 60.3 & 41.6 & 40.8 & 54.7 & 69.5 & 54.2 & 63.1 & 31.4 & 23.4 & 39.7 & 50.0 & 66.6 & 51.6 & 53.6 & 45.2 & 37.8 & 73.7 & 58.9 & 65.5 \\
\textbf{Qwen3.5} & 27B & \secondbest{58.0} & \secondbest{65.3} & \thirdbest{66.0} & \secondbest{66.4} & \best{65.9} & \secondbest{46.3} & \best{64.6} & \secondbest{73.2} & 53.4 & \thirdbest{68.9} & \secondbest{38.2} & \secondbest{27.6} & \secondbest{44.4} & \best{63.5} & \best{71.3} & \best{60.1} & \thirdbest{56.7} & \best{57.2} & \secondbest{44.9} & 77.1 & \thirdbest{63.3} & 77.0 \\
\textbf{Qwen3.5} & 35B-A3B & \best{58.1} & \best{66.5} & \best{69.0} & \thirdbest{64.6} & \thirdbest{60.7} & \thirdbest{45.3} & 56.6 & \secondbest{73.2} & \best{56.7} & \secondbest{69.1} & \thirdbest{38.0} & 27.1 & \best{45.8} & \secondbest{61.8} & \secondbest{71.0} & \thirdbest{55.3} & \best{59.2} & 50.3 & \thirdbest{42.5} & \secondbest{79.0} & 62.9 & \secondbest{92.9} \\
\midrule
\textbf{Qwen3.6} & 27B & \thirdbest{57.3} & 63.3 & 64.5 & \best{68.8} & \thirdbest{60.7} & \best{49.1} & \thirdbest{60.4} & \secondbest{73.2} & 50.8 & 68.7 & \best{39.2} & \best{28.3} & \thirdbest{42.5} & \thirdbest{59.6} & \thirdbest{70.6} & \secondbest{57.6} & 55.2 & \secondbest{54.7} & \best{45.0} & \thirdbest{78.3} & \secondbest{67.1} & 68.8 \\
\textbf{Qwen3.6} & 35B-A3B & 55.8 & 63.7 & \secondbest{66.6} & 62.6 & 51.8 & 44.6 & \secondbest{60.8} & \best{74.1} & \thirdbest{55.2} & \best{70.1} & 36.4 & 25.4 & \thirdbest{42.5} & 58.8 & \best{71.3} & 53.4 & \secondbest{58.0} & 52.1 & 39.7 & 78.1 & \secondbest{67.1} & \thirdbest{81.1} \\
\midrule
\textbf{InternVL3.5} & 4B & 32.1 & 41.2 & 44.5 & 27.6 & 13.4 & 28.3 & 40.6 & 55.0 & 32.7 & 45.5 & 9.5 & 11.7 & 23.2 & 27.2 & 59.5 & 34.2 & 37.6 & 32.5 & 13.6 & 65.2 & 41.1 & 46.2 \\
\textbf{InternVL3.5} & 8B & 32.9 & 43.6 & 45.1 & 26.5 & 13.6 & 28.1 & 42.5 & 54.3 & 36.1 & 46.9 & 8.6 & 12.4 & 24.3 & 33.5 & 59.1 & 37.8 & 37.0 & 33.9 & 12.8 & 67.9 & 41.5 & 50.3 \\
\textbf{InternVL3.5} & 14B & 29.9 & 37.8 & 41.7 & 24.5 & 11.8 & 24.8 & 37.3 & 53.4 & 31.9 & 42.7 & 9.5 & 12.0 & 19.8 & 26.4 & 55.8 & 28.2 & 35.4 & 29.5 & 9.2 & 67.6 & 40.0 & 45.9 \\
\midrule
\textbf{Molmo2} & 4B & 31.5 & 39.0 & 42.6 & 24.9 & 13.6 & 26.5 & 36.3 & 51.6 & 31.7 & 46.1 & 9.7 & 13.3 & 23.6 & 32.1 & 56.7 & 33.8 & 38.2 & 36.7 & 12.8 & 62.9 & 40.3 & 44.5 \\
\textbf{Molmo2} & 8B & 35.2 & 47.0 & 48.1 & 29.0 & 16.9 & 29.5 & 41.0 & 52.5 & 42.1 & 50.8 & 10.9 & 14.7 & 27.5 & 36.3 & 58.1 & 37.6 & 41.4 & 36.5 & 18.1 & 63.6 & 39.2 & 54.3 \\
\midrule
\textbf{Gemini-3.1} & -- & 44.8 & 48.8 & 51.8 & 51.8 & 45.7 & 31.1 & 42.9 & 57.5 & 48.5 & 58.7 & 28.8 & \thirdbest{27.3} & 31.1 & 45.6 & 54.1 & 44.3 & 46.6 & 47.0 & 33.2 & 70.0 & 41.6 & 56.3 \\
\textbf{GPT-5.5} & -- & 46.2 & 53.5 & 58.3 & 58.8 & 39.1 & 31.1 & 50.0 & 62.8 & 47.4 & 58.3 & 26.9 & 21.8 & 33.1 & 54.4 & 65.9 & 41.8 & 53.4 & \thirdbest{54.6} & 30.5 & \best{80.3} & 38.9 & 57.0 \\

\bottomrule
\end{tabular}
}}
\caption{
\textbf{Overall evaluation results on RoboStressBench.}
We report overall accuracy, performance across 16 fine-grained stress categories, and performance across five task dimensions. 
The best, second-best, and third-best results in each column are highlighted.
}
\label{tab:model_comparison}

\end{table*}

\begin{figure}[t]
    \centering
    \includegraphics[width=\linewidth]{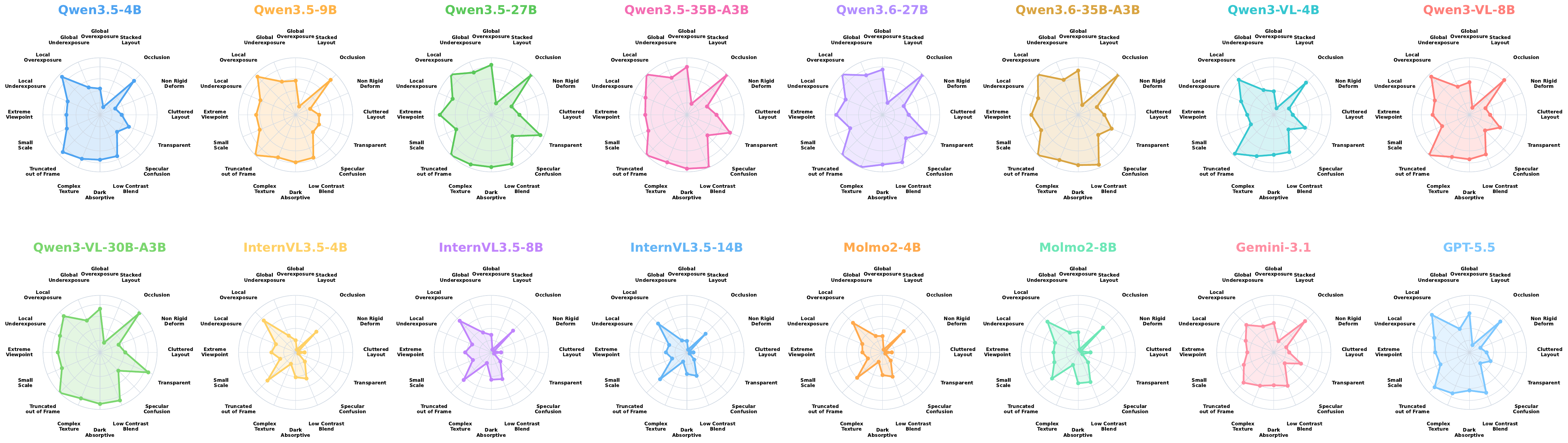}
    \caption{\textbf{Per-model dimension profiles on RoboStressBench.} Each panel shows one model’s scores over the 16 dimensions; see Table ~\ref{tab:model_comparison} for the raw numbers.
    }
    \label{fig:rader_small}

\end{figure}

\subsection{Main Benchmark Results}

Table~\ref{tab:model_comparison} reports the overall performance of all evaluated models on RoboStressBench. 
Fig.~\ref{fig:radar_large} and Fig.~\ref{fig:rader_small} further visualize model capabilities.

\paragraph{Takeaway 1: Physical visual stress remains challenging for current VLMs.}
Across all evaluated models, performance on RoboStressBench remains far from saturated. 
The best overall result is achieved by Qwen3.5-35B-A3B~\cite{qwen3.5} with only 58.1\% accuracy, while strong commercial models such as Gemini-3.1~\cite{comanici2025gemini} and GPT-5.5~\cite{achiam2023gpt} obtain 44.8\% and 46.2\%, respectively. 
These results indicate that current VLMs still struggle when recognition, reasoning, or planning depends on visually degraded evidence. 
Strong general-purpose visual understanding therefore does not necessarily translate into reliable performance under physically stressed scene conditions.

\paragraph{Takeaway 2: Scaling improves average performance but does not remove stress-specific weaknesses.}
Within the same model family, larger variants generally improve average performance, but the gains are uneven. 
For example, Qwen3.5~\cite{qwen3.5} improves from 49.8\% with the 4B model to 58.1\% with the 27B model, yielding an 8.3\% gain; Qwen3VL~\cite{bai2025qwen3} also improves from 43.2\% with the 4B model to 55.9\% with the 30B-A3B model. 
However, scaling does not consistently eliminate stress-specific failures: larger models still show low scores on the most challenging stress categories, and the InternVL3.5-14B~\cite{wang2025internvl3} variant even underperforms InternVL3.5-4B~\cite{wang2025internvl3} in overall accuracy. 
This suggests that physical stress introduces failure modes that are not fully resolved by increasing model scale alone.

\subsection{Stress-wise Analysis}
\begin{figure}[t]
    \centering
    \includegraphics[width=\linewidth]{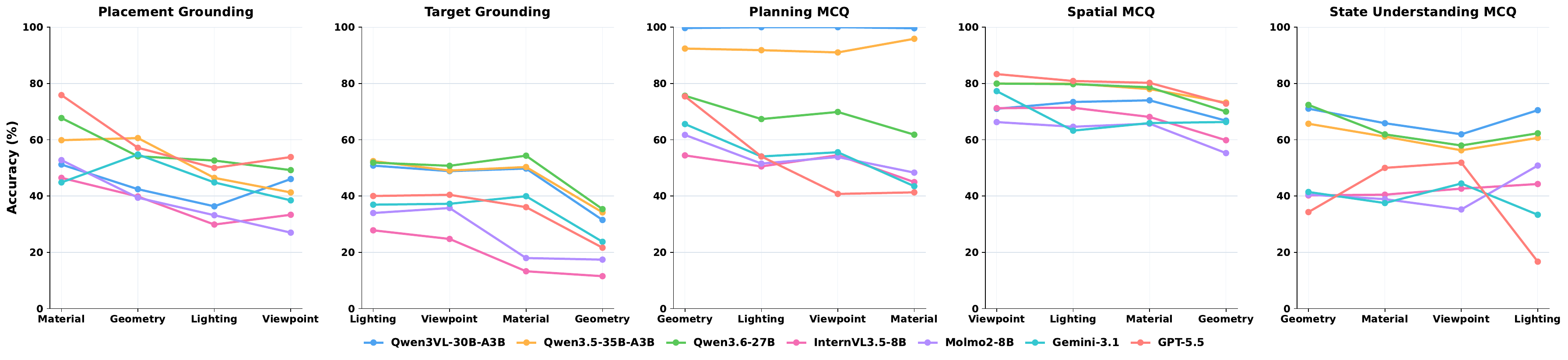}
    \caption{
    \textbf{Task-dependent sensitivity to physical visual stress.}
    For each task format, we visualize model accuracy across Material, Viewpoint, Lighting, and Geometry stress. 
    }
    \label{fig:lineplots}
\end{figure}

To identify where VLMs fail, we further break down performance across stress types. 
For each task, Fig.~\ref{fig:lineplots} plots model accuracy as a function of stress category.

\paragraph{Takeaway 3: Stress sensitivity is task-dependent.}
Physical stress affects VLM capabilities unevenly, and the dominant failure factor changes with the evaluated ability. 
As shown in Fig.~\ref{fig:lineplots}, Geometry stress is especially harmful for localization-oriented tasks: placement grounding, target grounding, and spatial MCQ generally reach their lowest accuracies under Geometry, suggesting that occlusion, clutter, and ambiguous spatial structure directly weaken object localization and spatial relation reasoning. 
In contrast, Planning MCQ does not follow the same Geometry-dominant pattern, several models remain relatively robust under Geometry but degrade more under Material or Viewpoint stress. 
State Understanding MCQ also shows a different profile, with noticeable drops under Lighting for some models. 
These results indicate that physical stress does not simply reduce overall image quality, but selectively disrupts different VLM capabilities depending on the task.

\subsection{StressDART Results}
\begin{wraptable}{r}{0.50\textwidth}
\vspace{-3mm}
\centering
\caption{\textbf{Results and ablation of StressDART.} We evaluate StressDART using Qwen3-VL-4B as the base model and compare different visual inputs to the final reasoner.}
\label{tab:stressdart}
\small
\begin{tabular}{lcc}
\toprule
\textbf{Method} & \textbf{Reasoner Input} & \textbf{Acc.} \\
\midrule
Qwen3-VL-4B & Original & 43.2\% \\
\midrule
StressDART & Rectified only & 48.9\% \\
StressDART & Original + Rectified & 49.0\% \\
\bottomrule
\end{tabular}
\vspace{-8mm}
\end{wraptable}
We next evaluate whether explicit stress diagnosis and targeted rectification can improve test-time reasoning. 
Using Qwen3-VL-4B~\cite{bai2025qwen3} as the base model, we report the results of StressDART in Table~\ref{tab:stressdart}. 
We also ablate the visual input to the final reasoner by comparing two settings: using only the rectified image, and using both the original and rectified images.

\paragraph{Takeaway 4: StressDART improves robustness through test-time rectification.}
StressDART improves over the Qwen3-VL-4B base model~\cite{bai2025qwen3} in both input settings, showing that explicit stress diagnosis and targeted rectification can help recover task-relevant visual evidence without updating model parameters. 
The rectified-only setting already provides most of the gain, suggesting that visual rectification is the main source of improvement. 
Providing both the original and rectified images yields the best accuracy, indicating that the original image can serve as a useful reference when visual editing introduces uncertainty or slightly changes local details. 
Overall, StressDART provides a practical test-time robustness improvement, while also pointing to future opportunities for more precise stress diagnosis and more content-preserving rectification.

\section{Conclusion}
We presented \textbf{RoboStressBench}, a physically grounded benchmark for evaluating VLM robustness under visual stress in embodied scenes. 
RoboStressBench organizes visual stress by four image-formation factors: Material, Viewpoint, Lighting, and Geometry. 
This design enables more interpretable diagnosis of model failures than treating degradation as arbitrary image corruption. 
The benchmark is built through human-annotated filtering, controlled stress synthesis, and real-world data collection. 
It covers diverse stress conditions across VQA and grounding tasks. 
Our evaluation of 16 VLMs shows that current models remain far from saturated under physical visual stress. 
It also shows that scaling alone does not eliminate stress-specific weaknesses. 
We further introduced \textbf{StressDART}, a test-time detect-and-rectify framework that improves robustness through stress diagnosis and targeted visual rectification. 
We hope RoboStressBench supports future research on VLMs that can perceive, reason, and act reliably under challenging real-world visual conditions.

\bibliographystyle{IEEEtran}
\bibliography{refs}


\appendix
\clearpage
\appendix
\section*{Appendix}

\section{RoboStressBench Details}
\label{supp:details}
\subsection{Data Sources}
RoboStressBench is constructed from three types of source data: existing public benchmarks, Internet-sourced real-world images, and self-collected images. 
For existing public benchmarks, we use samples from EmbSpatial-Bench~\cite{du2024embspatial}, RefSpatial-Bench~\cite{zhou2025roborefer}, RoboAfford-Eval~\cite{10.1145/3746027.3758209}, RoboSpatial-Home~\cite{song2025robospatial}, ManipulationVQA~\cite{chen2025robo2vlm}, VABench-P~\cite{yuan2026seeing}, Where2Place~\cite{yuan2025robopoint}, and RoboRefit~\cite{lu2023vlgrasp}. 
We retain the license and usage terms of each original source. 
For Internet-sourced examples, we mainly collect images from Pexels~\cite{pexels_2026} and follow the Pexels License, which allows free use and modification while restricting redistribution as standalone stock content. 
Self-collected images are captured by ourselves in diverse physical environments and are used to cover naturally occurring stress cases that are difficult to obtain from public datasets alone. 

\subsection{Annotation Protocol}
Six trained annotators with background knowledge in embodied AI and vision-language models perform the annotations.
For examples from existing public benchmarks, annotators first identify images that already exhibit physical visual stress and assign both coarse stress dimensions and fine-grained stress tags according to the Material, Viewpoint, Lighting, and Geometry taxonomy. 
When the original task annotations remain valid, we directly reuse the original questions and answers after manual verification. 
For images that do not originally contain clear stress but are suitable for controlled augmentation, we generate stressed variants and then manually check whether the original questions can still be reused. 
If a question becomes ambiguous or no longer matches the edited image, annotators revise the question or answer accordingly.

\begin{figure}[t]
    \centering
    \includegraphics[width=\linewidth]{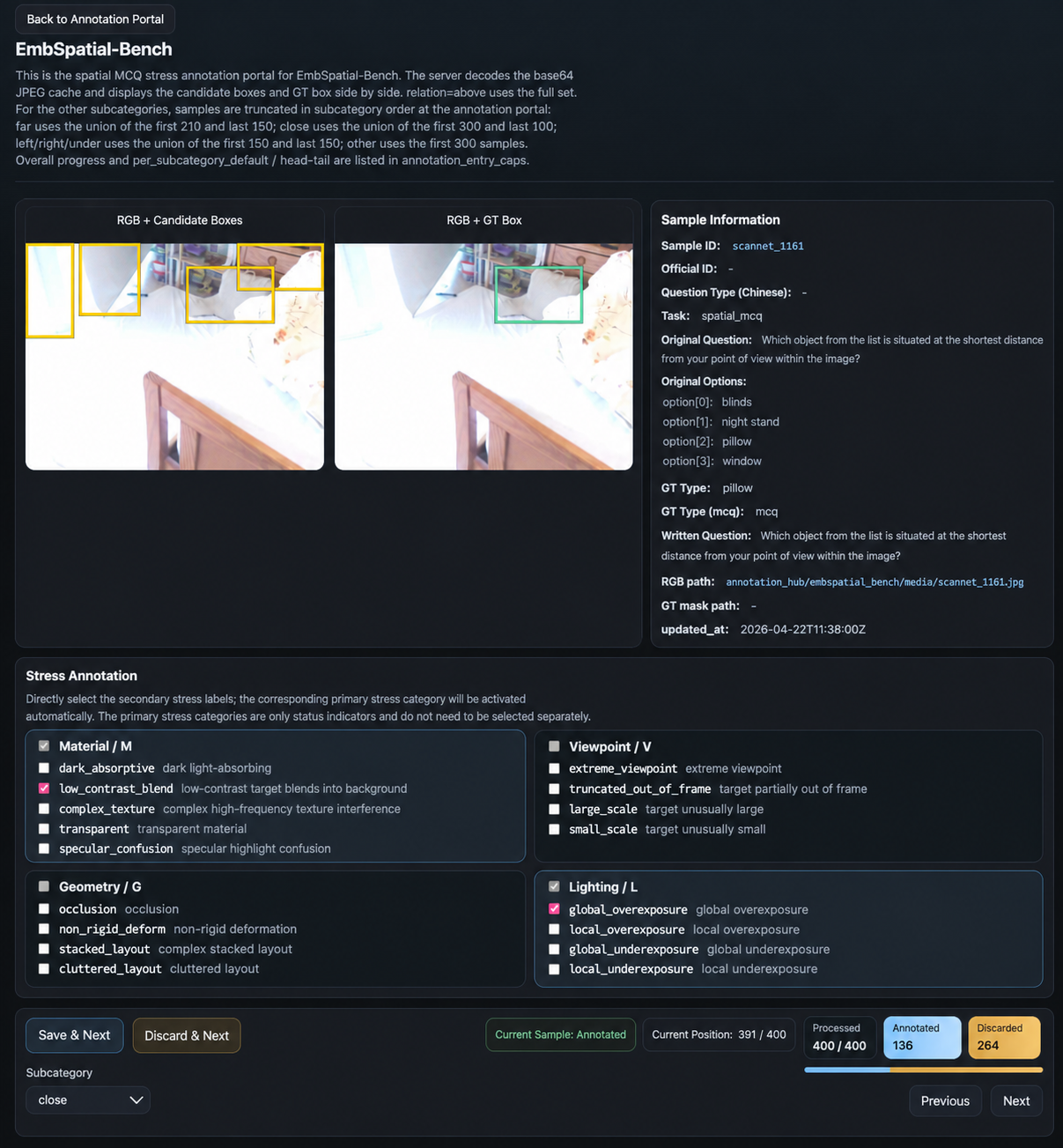}
    \caption{
    \textbf{Annotation interface for RoboStressBench.}
    Annotators inspect each image, assign coarse stress dimensions and fine-grained stress tags, verify or revise task questions and answers, and check grounding annotations when applicable.
    }
    \label{fig:annotation_interface}
\end{figure}

For Internet-sourced and self-collected images, we use a vocabulary-driven annotation pipeline. 
We first define a fixed object and scene vocabulary to guide candidate collection. 
Then, GroundingDINO~\cite{liu2024grounding} and SAM~\cite{kirillov2023segment} are used to generate candidate object annotations, which are ranked by confidence. 
Annotators manually inspect the candidates, remove noisy or ambiguous cases, assign stress labels, and write task-specific questions and answers. 
This process ensures that each example is physically meaningful, visually grounded, and aligned with the intended evaluation task.
The annotation interface is shown in Fig.~\ref{fig:annotation_interface}.

\subsection{Grounding Annotation Normalization}
For grounding tasks, we normalize all point and bounding-box annotations to a unified coordinate range of $[0,1000]$. 
During evaluation, models are explicitly prompted to output grounding results in the same normalized coordinate system. 
For point-based grounding, the model outputs a point coordinate $(x,y)$; for box-based grounding, it outputs a bounding box $(x_1,y_1,x_2,y_2)$, where all coordinates are represented in the $[0,1000]$ range.
This normalization makes grounding evaluation independent of the original image resolution and aspect ratio. 
It also allows different VLMs to follow a unified output format when images have different sizes, avoiding ambiguity caused by pixel-coordinate conventions.

\subsection{Controlled Stress Synthesis}

\label{app:controlled_stress_synthesis}

Controlled stress synthesis complements naturally occurring data by increasing coverage of stress categories that are rare, ambiguous, or difficult to isolate in real-world images. 
Starting from a nominal image, we generate a stressed counterpart by editing one intended physical factor from our taxonomy---Material, Viewpoint, Lighting, or Geometry---while preserving the task-relevant scene content. 
We use Gemini-3-Pro-Image~\cite{google2026geminiimage} and Qwen-Image-Edit~\cite{wu2025qwenimagetechnicalreport} as image editors, but treat them only as controlled perturbation tools: each edit prompt explicitly specifies the target stress category, the allowed visual change, and the scene elements that must remain unchanged.

Each synthesis job is defined by an \emph{edit profile}. 
An edit profile contains: 
(i) a nominal image and its original task annotation; 
(ii) a target stress category; 
(iii) an editing instruction describing the desired physical change; 
(iv) preservation constraints for task-relevant objects, layout, camera pose, lighting consistency, and photorealism; and 
(v) an annotation policy specifying whether the original label can be reused. 
For grounding tasks, we resize the nominal and stressed images to the same resolution and check whether the target region remains pixel-aligned. 
If alignment is preserved, we transfer the original normalized point or bounding box annotation. 
If the edit changes the target location, shape, visibility, or surrounding evidence, annotators re-label the stressed image. 
All generated samples are manually verified for three conditions: the intended stress is visually present, the task semantics remain valid, and the final annotation is correct.

\begin{figure}[t]
  \centering
  \includegraphics[width=\linewidth]{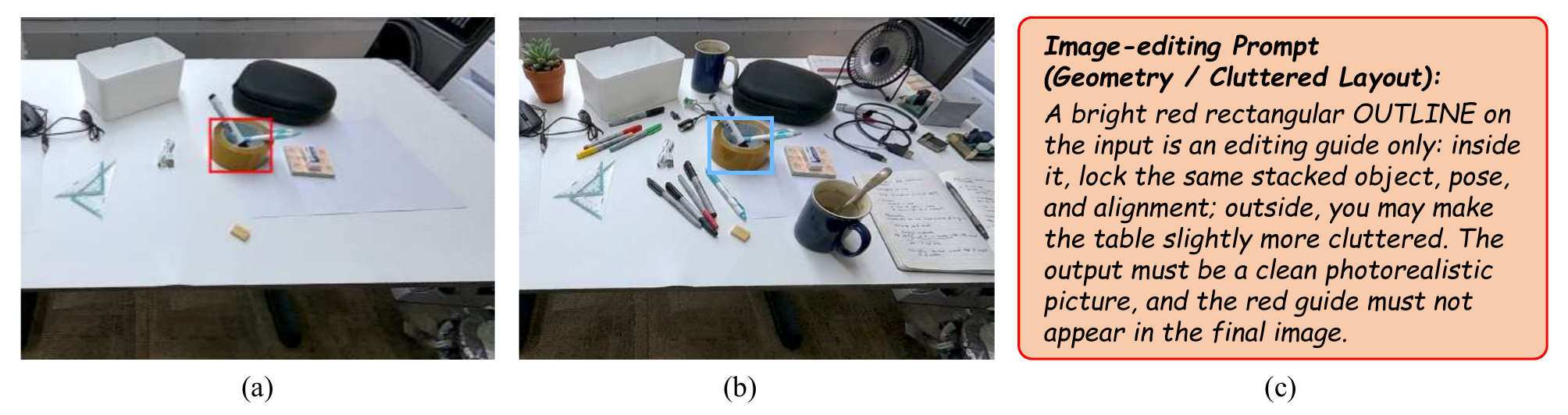}
  \caption{\textbf{Controlled synthesis with a temporary bounding-box guide.}
  (a) Nominal inputs with rasterized red rectangles used only as editing guides. 
  The guides indicate regions whose target object, pose, and alignment should be preserved during editing. 
  (b) Stressed outputs after adding surrounding clutter; cyan boxes visualize annotations that are reused only when post-edit alignment is verified. 
  (c) Example edit prompt specifying the target stress, the protected region, and the requirement that the guide must not appear in the final image.}
  \label{fig:supp_geom_synth_cluttered}
\end{figure}

\begin{figure}[t]
  \centering
  \includegraphics[width=\linewidth]{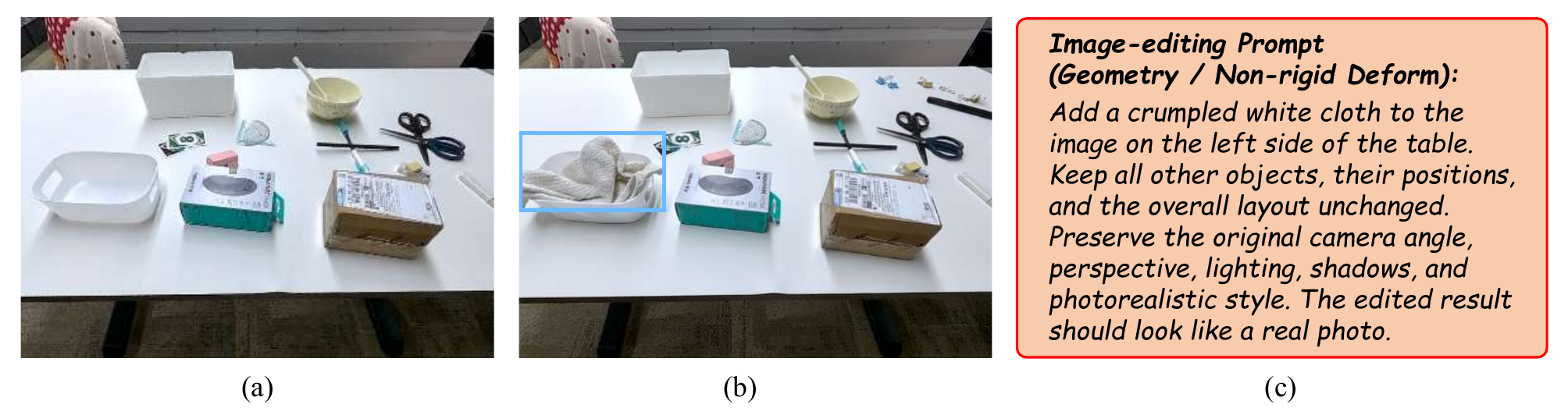}
  \caption{\textbf{Language-only synthesis for spatial and non-rigid stress.}
  (a) Nominal scene without rasterized guides. 
  (b) Stressed output with an inserted deformable foreground object; boxes indicate annotator-verified regions after editing. 
  (c) Example prompt that specifies the non-rigid object, its placement, and preservation constraints for other objects, camera pose, lighting, and photorealistic style.}
  \label{fig:supp_geom_synth_nonrigid}
\end{figure}

\begin{figure}[t]
  \centering
  \includegraphics[width=\linewidth]{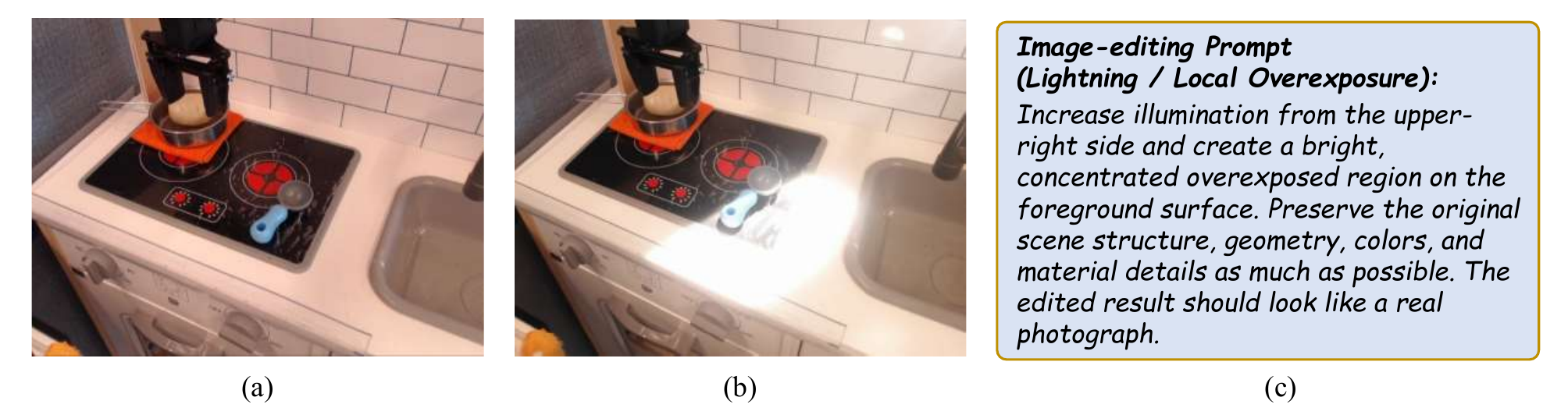}
  \caption{\textbf{Appearance-factor synthesis via local overexposure.}
  (a) Nominal scene before editing. 
  (b) Stressed output with a bright, localized overexposed region produced by a directional lighting change, while the overall scene structure and task-relevant objects are preserved where possible. 
  (c) Example prompt describing the lighting manipulation and preservation constraints.}
  \label{fig:supp_light_synth_local_overex}
\end{figure}

\begin{figure}[t]
  \centering
  \includegraphics[width=\linewidth]{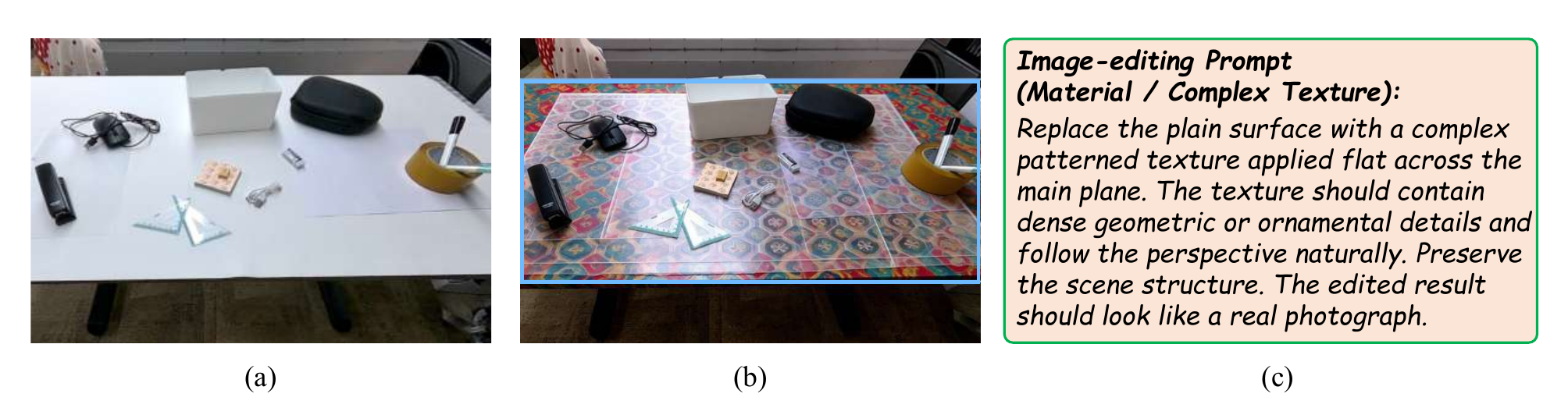}
  \caption{\textbf{Appearance-factor synthesis via complex texture.}
  (a) Nominal scene before editing. 
  (b) Stressed output with dense texture or typographic patterns applied to a surface in a perspective- and lighting-consistent manner, while keeping the manipulation layout stable when possible. 
  (c) Example prompt describing the material change and the requirement to preserve scene structure and photorealism.}
  \label{fig:supp_mat_synth_comp_text}
\end{figure}

Figures~\ref{fig:supp_geom_synth_cluttered}--\ref{fig:supp_mat_synth_comp_text} illustrate representative synthesis protocols. 
We organize these examples by \emph{how the edit is controlled}, rather than by enumerating every stress axis or sub-category. 
The selected cases cover three common control modes used throughout the pipeline: temporary spatial guides, language-only spatial edits, and appearance-factor edits. 
Across all modes, the principle is the same: introduce a controlled physical stress, preserve task-relevant semantics, and decide annotation reuse only after post-edit validation.

\paragraph{Region-guided preservation.}
Figure~\ref{fig:supp_geom_synth_cluttered} illustrates a guide-based protocol for cases where the target annotation should remain stable while the surrounding scene becomes more challenging. 
We rasterize the original bounding box as a temporary editing guide on the reference image and instruct the editor to keep the object inside the guide fixed while adding clutter outside the guide. 
The guide is used only during synthesis and is never included in the final evaluation image. 
This protocol is useful when the desired stress affects the nearby spatial context, such as clutter or background complexity, rather than the target object itself. 
When the edited image remains aligned with the nominal image, the original grounding annotation and question can be reused.

\paragraph{Language-only spatial edits.}
Figure~\ref{fig:supp_geom_synth_nonrigid} shows a complementary protocol for edits that are easier to specify with natural language than with a rasterized guide. 
The prompt describes the inserted or modified object, its spatial placement, and the scene elements that must remain unchanged. 
This mode is useful for non-rigid geometry stress, deformable foreground insertions, and other cases where exact pixel preservation is not assumed. 
Because the inserted object may change the visible scene layout or object extent, annotators verify the edited content and update grounding labels whenever the target region, visibility, or spatial evidence changes.

\paragraph{Appearance-factor edits.}
Figures~\ref{fig:supp_light_synth_local_overex} and~\ref{fig:supp_mat_synth_comp_text} illustrate controlled edits to appearance factors such as illumination and surface material. 
These edits do not aim to change the object layout; instead, they modify visual evidence by altering how the existing scene is seen. 
For lighting stress, the prompt can introduce a concentrated highlight, glare region, shadow pocket, or exposure change while preserving global geometry and object identity. 
For material stress, the prompt can replace a plain surface with a dense texture, decal, or typographic pattern that follows the scene perspective and lighting. 
In both cases, annotators check that the edited appearance is physically plausible, that the intended stress category is satisfied, and that task-relevant evidence remains valid. 
Grounding annotations are reused only when the target remains aligned; otherwise, the stressed image is re-annotated.

\paragraph{Scope of examples.}
The examples above are representative synthesis profiles, not an exhaustive catalog of all fine-grained stress categories. 
Many sub-categories follow the same workflow with different prompt instantiations: specify the target stress, preserve task-relevant content, generate the stressed variant, and verify annotation validity. 
The selected examples cover the main control modes in our pipeline---region-guided preservation, language-only spatial editing, and appearance-factor editing---and therefore illustrate the full synthesis and quality-control procedure.

\subsection{Dataset Statistics}
RoboStressBench contains 7183 examples in total. 
Among them, 2927 examples are filtered from existing unconstrained datasets, 2596 examples are generated through controlled stress synthesis, and 1660 examples are collected from additional real-world sources, including Internet-sourced images and images captured by ourselves. 
This combination allows the benchmark to include both naturally occurring stress cases and controlled high-stress variants.

In terms of stress distribution, RoboStressBench includes 2785 Material examples, 1292 Viewpoint examples, 1753 Lighting examples, and 3327 Geometry examples. 
For Material stress, the dataset contains 711 dark absorptive, 761 low-contrast blend, 551 complex texture, 575 transparent, and 495 specular-confusion examples. 
For Viewpoint stress, it contains 212 extreme-viewpoint, 665 truncated-out-of-frame, and 496 small-scale examples. 
For Lighting stress, it contains 364 global-overexposure, 575 local-overexposure, 521 global-underexposure, and 319 local-underexposure examples. 
For Geometry stress, it contains 1205 occlusion, 579 non-rigid-deformation, 865 stacked-layout, and 1658 cluttered-layout examples.
Note that a single example may be associated with multiple stress tags; consequently, the per-tag counts are reported independently and their sum may exceed the total number of examples.

RoboStressBench also covers multiple evaluation tasks. 
Specifically, it contains 949 placement-grounding examples, 3411 target-grounding examples, 1369 spatial-reasoning multiple-choice examples, 633 state-understanding multiple-choice examples, and 821 planning multiple-choice examples. 
These task types are designed to evaluate complementary embodied capabilities, including object localization, target grounding, spatial relation reasoning, object-state understanding, and high-level planning under physical visual stress.
We provide representative examples from RoboStressBench in Fig.~\ref{fig:case_1}--Fig.~\ref{fig:case_5}.
All examples and annotations of RoboStressBench are provided in the supplementary material.

\begin{figure}[p]
  \centering
  \includegraphics[width=\linewidth]{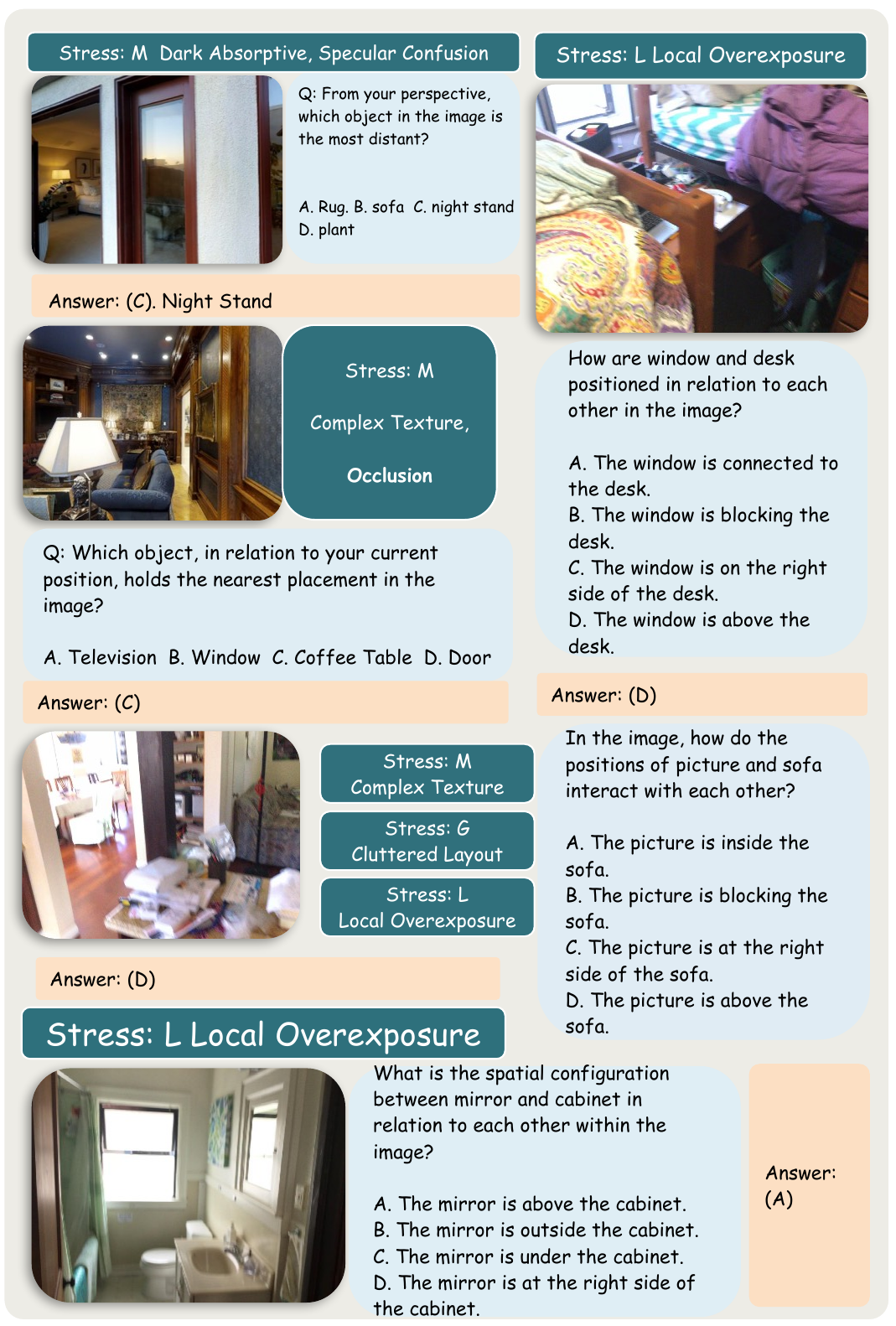}
  \caption{\textbf{Representative examples from RoboStressBench.}We show several physically stressed examples with their questions, answers, and stress annotations.
  }

  \label{fig:case_1}
\end{figure}

\begin{figure}[p]
  \centering
  \includegraphics[width=\linewidth]{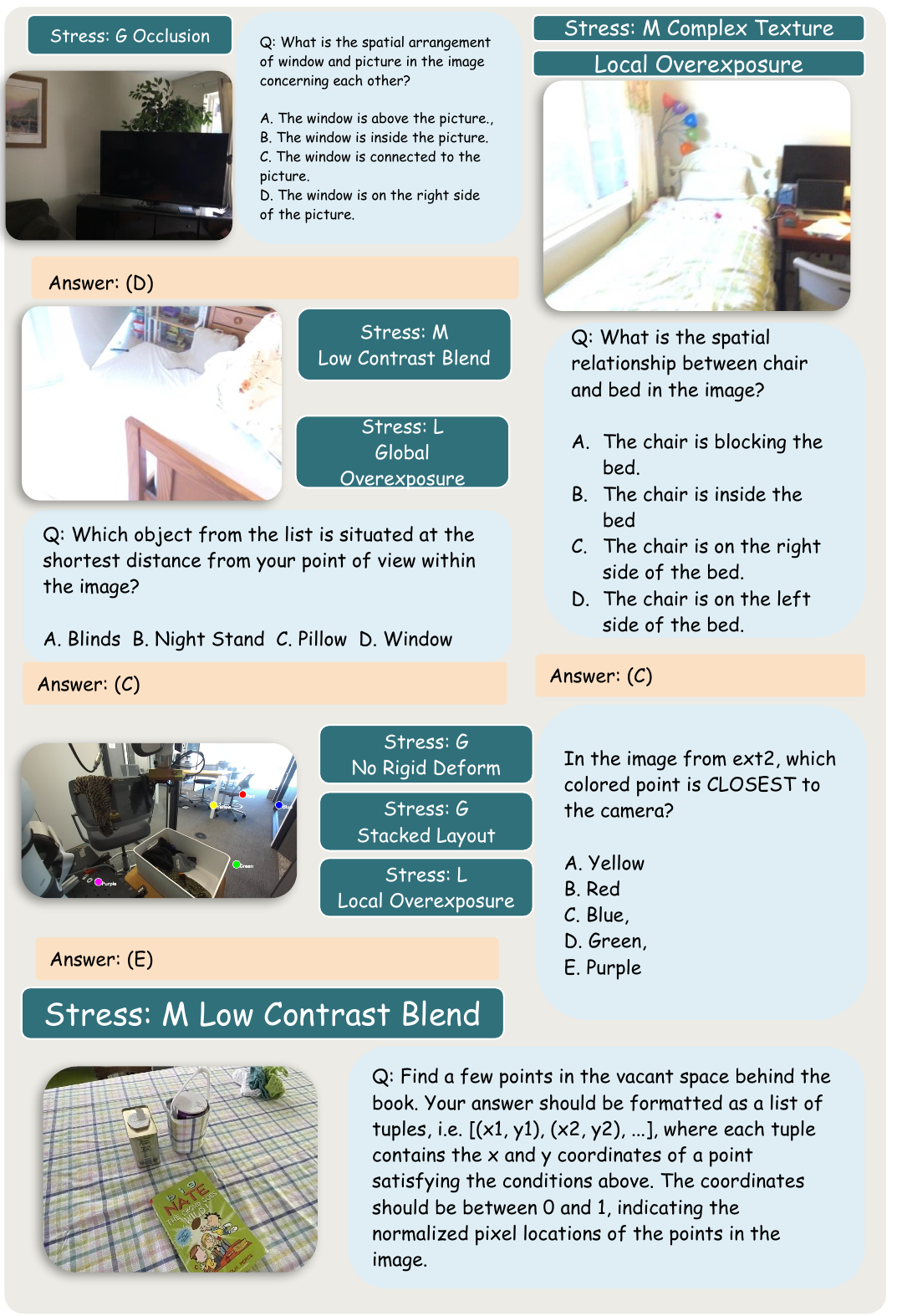}
  \caption{\textbf{Representative examples from RoboStressBench.}We show several physically stressed examples with their questions, answers, and stress annotations.
  }
  \label{fig:case_2}
\end{figure}

\begin{figure}[p]
  \centering
  \includegraphics[width=\linewidth]{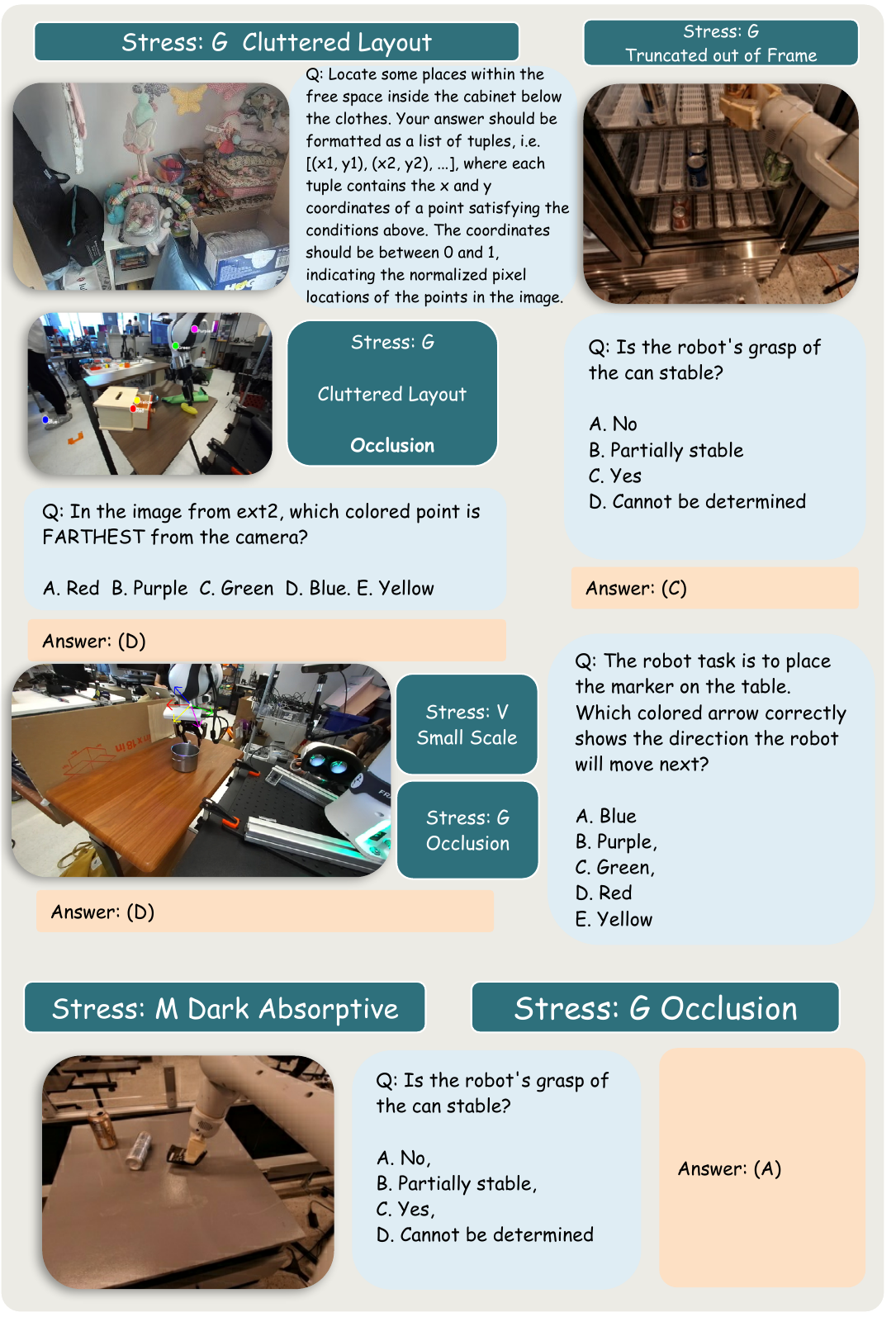}
  \caption{\textbf{Representative examples from RoboStressBench.}We show several physically stressed examples with their questions, answers, and stress annotations.
  }
  \label{fig:case_3}
\end{figure}

\begin{figure}[p]
  \centering
  \includegraphics[width=\linewidth]{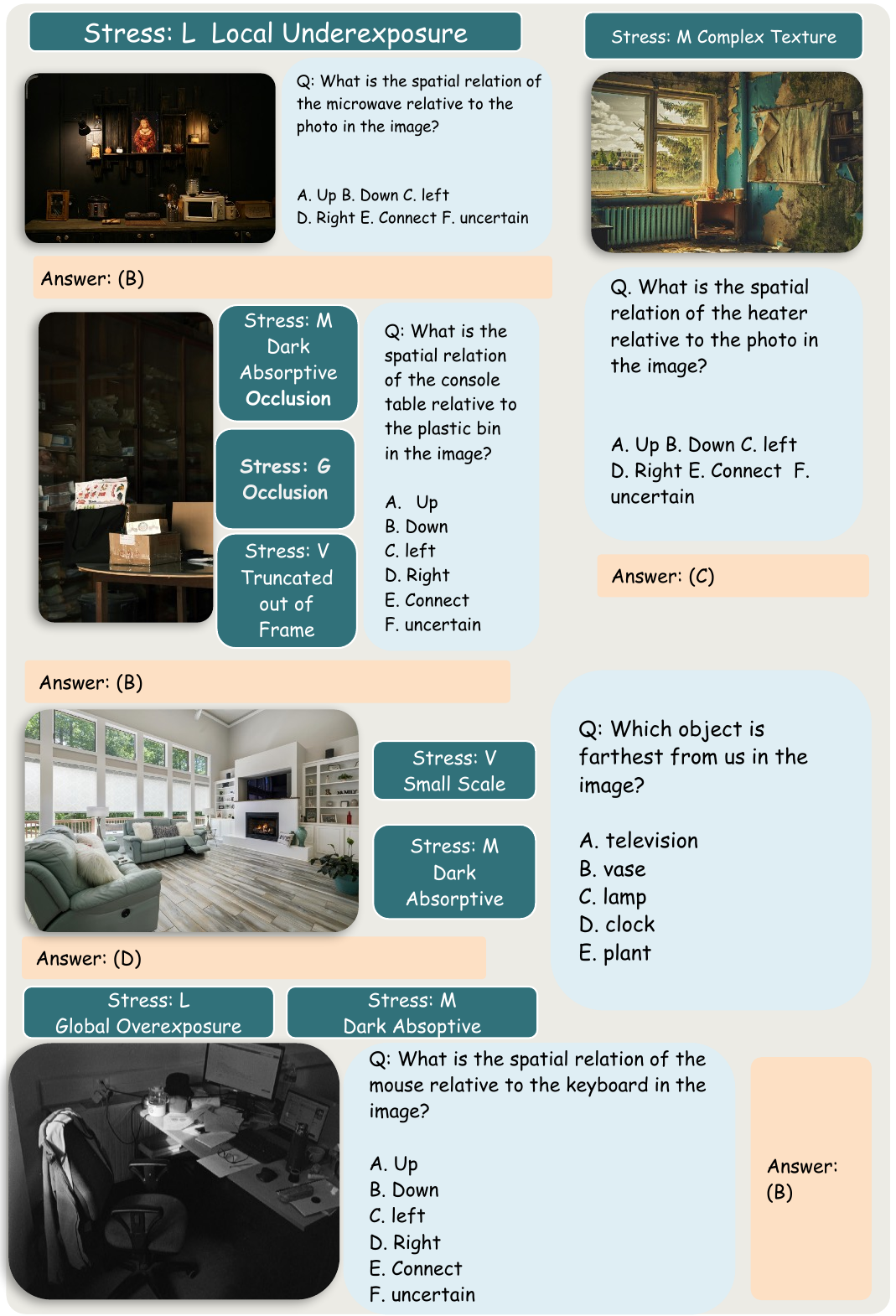}
  \caption{\textbf{Representative examples from RoboStressBench.}We show several physically stressed examples with their questions, answers, and stress annotations.
  }
  \label{fig:case_4}
\end{figure}

\begin{figure}[p]
  \centering
  \includegraphics[width=\linewidth]{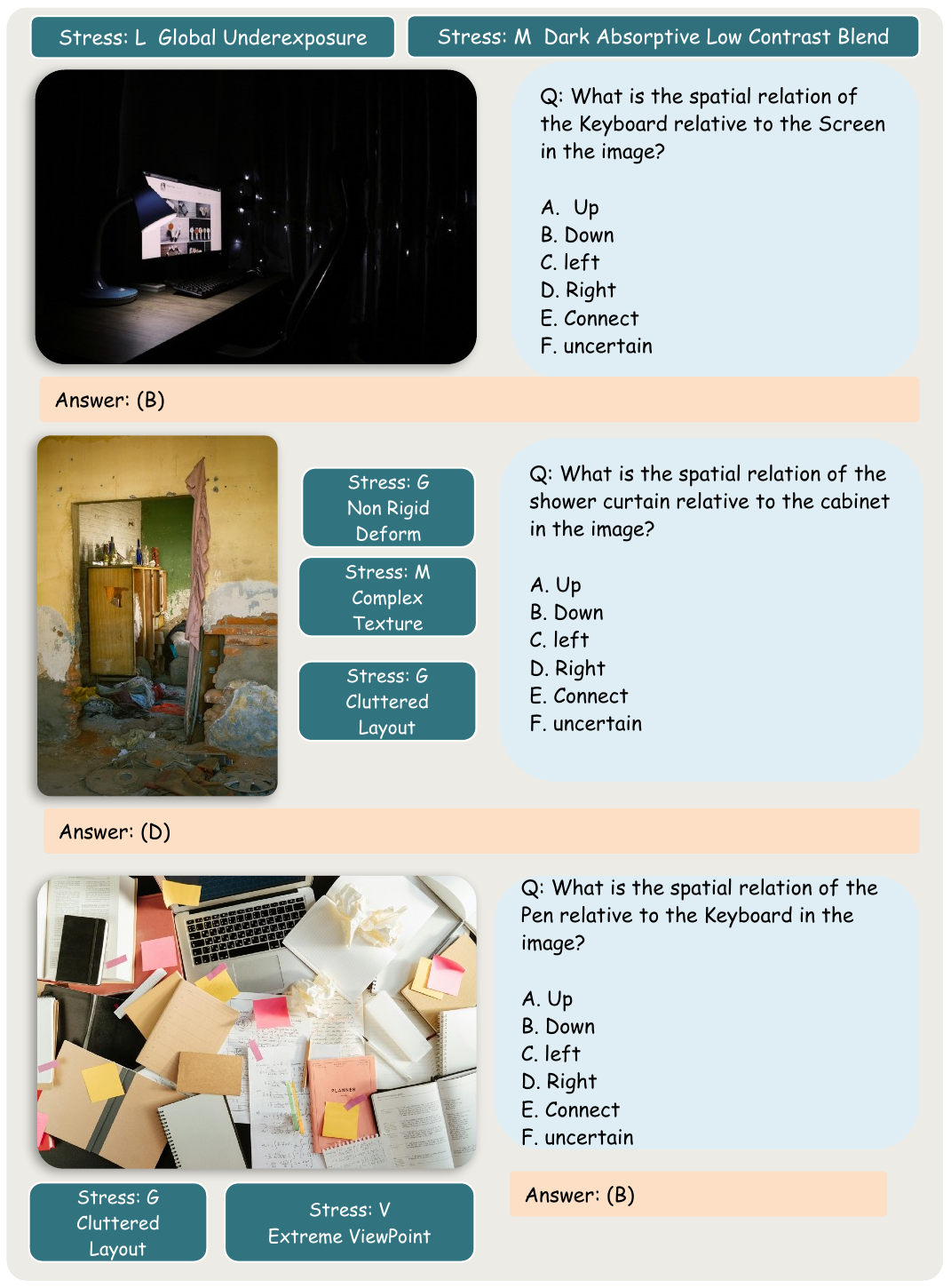}
  \caption{\textbf{Representative examples from RoboStressBench.}We show several physically stressed examples with their questions, answers, and stress annotations.
  }
  \label{fig:case_5}
\end{figure}

\section{Additional Experimental Details}
\label{supp:exp_details}

\subsection{Detailed Grounding Results}
\label{supp:grounding_metrics}

RoboStressBench contains both point-based and box-based grounding tasks. 
For point-based grounding, a prediction is considered correct if the predicted point falls inside the ground-truth mask. 
For box-based grounding, we follow COCO-style IoU evaluation and report IoU@0.50, IoU@0.95, and the mean accuracy averaged over thresholds from 0.50 to 0.95 with a step size of 0.05. 
In Table~\ref{tab:model_comparison}, the grounding score is computed as the average of point-based grounding accuracy and box-based IoU@0.95. 
Table~\ref{tab:box_grounding_metrics} reports the separate point-based results and the complete box-based grounding metrics.

\begin{table*}[t]
\centering
\caption{
\textbf{Detailed grounding results.}
We report point-based grounding accuracy and box-based grounding metrics, including IoU@0.50, IoU@0.95, and mean accuracy averaged over IoU thresholds from 0.50 to 0.95 with a step of 0.05, following the standard COCO-style protocol.
}
\label{tab:box_grounding_metrics}
{\setlength{\tabcolsep}{7pt}
\renewcommand{\arraystretch}{1.15}
\small
\resizebox{\textwidth}{!}{
\begin{tabular}{llcccc}
\toprule
\textbf{Model} & \textbf{Size} & \textbf{Point-acc} & \textbf{IoU@0.50} & \textbf{IoU@0.95} & \textbf{mAcc} \\
\midrule
Qwen3VL & 4B & 50.1 & 80.4 & 26.6 & 68.6 \\
Qwen3VL & 8B & 54.4 & 79.7 & 24.7 & 65.8 \\
Qwen3VL & 30B-A3B & 53.5 & 82.0 & 30.4 & 70.0 \\
Qwen3.5 & 4B & 52.0 & 82.0 & 25.7 & 67.7 \\
Qwen3.5 & 9B & 56.9 & \thirdbest{82.6} & 25.5 & 68.8 \\
Qwen3.5 & 27B & \best{64.3} & \secondbest{83.1} & \secondbest{34.3} & \best{72.4} \\
Qwen3.5 & 35B-A3B & 58.9 & 82.0 & \thirdbest{32.1} & \thirdbest{70.5} \\
Qwen3.6 & 27B & \secondbest{63.3} & \best{83.3} & \best{34.5} & \best{72.4} \\
Qwen3.6 & 35B-A3B & 59.9 & 80.6 & 28.0 & 68.7 \\
InternVL3.5 & 4B & 37.7 & 18.8 & 0.5 & 8.4 \\
InternVL3.5 & 8B & 37.1 & 27.6 & 0.5 & 11.0 \\
InternVL3.5 & 14B & 28.0 & 7.9 & 0.0 & 2.2 \\
Molmo2 & 4B & 38.2 & 3.4 & 0.0 & 0.9 \\
Molmo2 & 8B & 49.5 & 4.4 & 0.1 & 1.3 \\
Gemini-3.1 & -- & 58.3 & 60.3 & 18.8 & 45.3 \\
GPT-5.5 & -- & \thirdbest{60.4} & 80.3 & 15.0 & 61.2 \\
\bottomrule
\end{tabular}
}}
\end{table*}

\subsection{Compute Resources}
\label{supp:cost}

All experiments in RoboStressBench are conducted in an inference-only setting, without model fine-tuning or parameter updates. 
For open-source VLMs, we run evaluation on 8 NVIDIA H100 GPUs, each with 80 GB memory, using the official inference implementations of each model. 
All models are evaluated with deterministic greedy decoding, setting the maximum generation length to 64 new tokens and disabling sampling (temperature = 0.0, top-p = 1.0), as described in Sec.~\ref{sec:exp}. 
The full open-source model evaluation takes approximately 48 GPU-hours in total.

For StressDART, the base reasoner is Qwen3-VL-4B~\cite{bai2025qwen3}, and the Stress Rectifier is implemented with Qwen-Image-Edit~\cite{wu2025qwenimagetechnicalreport}. 
This rectification step introduces additional test-time image-editing cost, requiring approximately 150 GPU-hours for the evaluated subset. 
Closed-source models, including Gemini-3.1~\cite{comanici2025gemini} and GPT-5.5~\cite{achiam2023gpt}, are evaluated through official APIs and therefore do not consume local GPU resources.

\section{Limitations}
\label{supp:limitations}
RoboStressBench is designed as a diagnostic benchmark for physical visual stress in embodied scenes, but it still has several limitations. First, although our Material--Viewpoint--Lighting--Geometry taxonomy is physically grounded and interpretable, it is not intended to exhaust all possible sources of visual difficulty in real-world embodied environments. Although RoboStressBench supports multi-label stress annotation, stress axes are not perfectly orthogonal in real scenes; factors such as viewpoint and geometry or lighting and material appearance can still be entangled, making fine-grained attribution challenging.

Second, our dataset construction combines human-curated filtering, controlled stress synthesis, and additional real-world collection. While this design balances realism, diversity, and controllability, it may still introduce source bias from the datasets and scenes we sample, as well as artifacts from generative editing for synthesized stress cases. We reduce this risk through manual verification and re-annotation when necessary, but synthetic examples cannot fully replace naturally occurring physical stress.

Third, the current benchmark focuses on image-based VQA and grounding tasks. These tasks capture important perception, spatial reasoning, and planning-related abilities, but they do not fully evaluate closed-loop embodied behavior, long-horizon interaction, or temporal robustness in dynamic scenes. Extending RoboStressBench to video observations, multi-view interaction, and real robot execution would provide a more complete picture of embodied robustness. 

Finally, StressDART is an initial test-time intervention rather than a fully optimized robustness framework. 
Its results show that explicit stress diagnosis and targeted rectification can substantially improve performance. 
Nevertheless, some negative flips still occur when visual editing changes task-relevant cues or when the diagnosed stress does not match the true failure mode. 
Future work should investigate more reliable stress detectors, content-preserving rectification methods, and reasoning strategies that can better decide when to trust the original image, the rectified image, or both. 
We hope these limitations will motivate future research on more realistic, temporally grounded, and action-aware robustness evaluation for embodied VLMs.

\section{Broader Impacts}
\label{supp:impacts}
RoboStressBench aims to support the development of more reliable VLMs for embodied AI by exposing failures under physically plausible visual stress. 
Such evaluation can benefit robotics systems that must operate in challenging real-world environments, including low illumination, occlusion, reflective materials, unusual viewpoints, and cluttered scenes. 
By providing stress annotations and task-level evaluation, RoboStressBench can help researchers diagnose when VLM perception is unreliable and develop targeted robustness improvements before deployment.

At the same time, the benchmark and StressDART inherit broader risks associated with VLM-based embodied systems. 
Models may still hallucinate answers, mislocalize objects, or overestimate their confidence under severe visual ambiguity. 
When integrated into robotic pipelines, such errors may lead to unsafe manipulation, navigation, or planning decisions, especially in high-stakes environments. 
StressDART can improve robustness through test-time rectification, but visual editing may also alter task-relevant evidence if applied incorrectly, so it should not be treated as a substitute for calibrated uncertainty estimation or safety checks.

We believe that releasing RoboStressBench to the research community can have positive impact by enabling more systematic evaluation of perception robustness under realistic physical stress. 
Open access to the benchmark, annotations, and evaluation protocol can facilitate reproducible comparison, encourage stress-aware model development, and support safer embodied AI systems across robotic platforms such as mobile robots, robotic arms, and humanoids.

\section{Impact Mitigation Measures}
RoboStressBench is an evaluation benchmark, not a deployed embodied agent, but we still consider the possible risks related to data release and benchmark usage. We will document the dataset sources, stress taxonomy, annotation process, task formats, evaluation metrics, and inference settings to make the benchmark transparent and reproducible. When releasing the data, we will follow the licenses of the original sources, remove or exclude images with personally identifiable or sensitive information, and clearly indicate which samples are real and which are synthesized. We will also release the benchmark data, annotation schema, evaluation scripts, and usage instructions as soon as possible to support responsible use by the community.

We will also provide clear guidance on what the benchmark should and should not be used for. RoboStressBench is intended to help researchers diagnose how VLMs fail under realistic physical visual stress, rather than to support surveillance, biometric identification, or high-stakes automated decisions. Similarly, StressDART should be viewed as an exploratory test-time strategy, not as a complete safety mechanism, since visual editing may sometimes change important visual cues. Therefore, we encourage users to report both successful and failed cases, keep the original image available during reasoning, and treat benchmark results as evidence for improving model robustness rather than as proof that a model is ready for real-world deployment.

\section{Licenses}
\label{supp:licenses}

RoboStressBench is constructed from existing public benchmarks, Pexels-sourced real-world images, and controlled stress synthesis. We retain the license and usage terms of each original data source. Our annotations, metadata, and benchmark construction code may be released under our chosen research license, while images and derived visual assets remain subject to the licenses or terms of their corresponding source data.

\begin{enumerate}
    \item \textbf{Existing public benchmarks.} RoboStressBench uses samples from EmbSpatial-Bench~\cite{du2024embspatial}, released under CC BY 4.0; RefSpatial-Bench~\cite{zhou2025roborefer}, released under Apache 2.0; RoboAfford-Eval~\cite{10.1145/3746027.3758209}, released under CC BY 4.0; RoboSpatial-Home~\cite{song2025robospatial}, released under Apache 2.0; ManipulationVQA~\cite{chen2025robo2vlm}, released under Apache 2.0; VABench-P~\cite{yuan2026seeing}, released under Apache 2.0; and Where2Place~\cite{yuan2025robopoint}, released under Apache 2.0. RoboRefit~\cite{lu2023vlgrasp}  is distributed via the official VL-Grasp repository without an explicit dataset license; we use it for non-commercial academic research only, consistent with common practice for unlicensed academic datasets.

    \item \textbf{Pexels-sourced real-world images.} The dataset contains images sourced from Pexels~\cite{pexels_2026}. Under the Pexels License, content is free to use and modify for commercial or non-commercial purposes without required attribution. The terms explicitly prohibit redistributing or selling the photos on other stock photo or wallpaper platforms. We release these images exclusively as part of an academic benchmark dataset, which strictly complies with these terms. Users of our benchmark are also subject to the original Pexels License.
    

    \item \textbf{Controlled stress synthesis.} Some controlled stress samples are synthesized from existing benchmark images, such as lighting-stress variants generated from public benchmark sources. These derived samples inherit the license and usage constraints of their underlying source datasets and are not relicensed independently. Synthesis based on proprietary in-house data uses raw images provided by an industrial partner. We plan to release these specific derived samples under a research-only, non-commercial license to protect the proprietary nature of the original assets.
    
\end{enumerate}



\end{document}